\documentclass[11pt, a4paper]{dm}

\usepackage{amsmath,amssymb,amsfonts}
\usepackage{mathrsfs}
\usepackage[super,sort&compress]{natbib}
\setcitestyle{citesep={\textcolor{blue}{,}}}

\usepackage{multirow}
\usepackage{booktabs}
\usepackage{algorithm}
\usepackage{algorithmicx}
\usepackage{algpseudocode}
\usepackage{listings}
\usepackage{tabularray}
\usepackage{pdflscape}
\usepackage[title]{appendix}
\usepackage{longtable}
\usepackage{array}

\usepackage{graphicx}
\usepackage{xcolor}
\usepackage{textcomp}
\usepackage{manyfoot}
\usepackage{hyperref}
\usepackage{lineno}         
\usepackage{xspace}
\usepackage{mdframed}
\usepackage{float}          
\usepackage{fltpage}
\usepackage{language}

\usepackage[most]{tcolorbox}

\DeclareCaptionLabelSeparator{nrcpipe}{ $\vert$ }
\makeatletter
\renewcommand\AB@affilsepx{\protect\\[\affilsep]\protect\Affilfont}
\newlength{\authoraffilgap}
\renewcommand\@author{\ifx\AB@affillist\AB@empty\AB@author\else
      \ifnum\value{affil}>\value{Maxaffil}\def\rlap##1{##1}%
    \AB@authlist\\[\authoraffilgap]\AB@affillist
    \else \AB@authors\fi\fi}
\makeatother

\newcommand{\clinmm}{ClinMM-Bench\xspace}

\definecolor{Stone0}{HTML}{EFF3F9}
\definecolor{Stone1}{HTML}{DEE5F0}
\definecolor{Stone2}{HTML}{5F82AD}
\newenvironment{pboxed}[1][]{
    \begin{tcolorbox}[
        enhanced,
        breakable,
        colback=Stone0,
        frame hidden,
        boxrule=0pt,
        borderline west={10pt}{0mm}{Stone2},
        boxsep=2pt,
        left=12pt,
        right=2pt,
        top=2pt,
        bottom=2pt,
        before upper={
            \setlength{\parskip}{5pt}
            \small
        },
        #1
    ]
}
{%
    \par%
    \end{tcolorbox}%
}
\makeatletter
\let\AbstractContent\abscontent
\renewcommand{\abscontent}{}
\makeatother

\title{\centering
Evaluating Multi-Turn Multimodal Diagnostic Reasoning \\ 
on Challenging Real-World Clinical Cases}

\author[1,2,3,4]{Rui Yang}
\author[5,6]{Weihao Xuan}
\author[3]{Yi Lin}
\author[7]{Zhuhan Bao}
\author[8]{Jonathan Chong Kai Liew}
\author[9]{Matthew Yu Heng Wong}
\author[8,10]{Nicol\'{a}s Lescano}
\author[8,10,11]{Nikita R. Paripati}
\author[12,13]{Emily Ling-Lin Pai}
\author[14]{Jiarui Liu}
\author[6]{Heli Qi}
\author[15]{Heng-Jui Chang}
\author[16,17]{Benny Kai Guo Loo}
\author[1,2]{Huitao Li}
\author[1,2]{Kunyu Yu}
\author[3]{Yufan Wang}
\author[7]{Chuan Hong}
\author[18]{Shijian Lu}
\author[19]{Douglas Teodoro}
\author[5,6]{Naoto Yokoya}
\author[8,20]{Ross Koppel}
\author[14]{Mona Diab}
\author[21]{Hua Xu}
\author[22,23,24]{David W. Bates}
\author[1,2,7,25,26$\dagger$]{Nan Liu}
\author[3,27$\dagger$]{Yifan Peng}

\affil[1]{Center for Biomedical Data Science, Duke-NUS Medical School,
Singapore, Singapore}
\affil[2]{Duke-NUS AI + Medical Sciences Initiative, Duke-NUS Medical
School, Singapore, Singapore}
\affil[3]{Department of Population Health Sciences, Weill Cornell
Medicine, New York, NY, USA}
\affil[4]{System Engineering, College of Engineering, Cornell
University, Ithaca, NY, USA}
\affil[5]{Graduate School of Frontier Sciences, The University of
Tokyo, Chiba, Japan}
\affil[6]{RIKEN Center for Advanced Intelligence Project, Tokyo, Japan}
\affil[7]{Department of Biostatistics and Bioinformatics, Duke
University, Durham, NC, USA}
\affil[8]{Perelman School of Medicine, University of Pennsylvania,
Philadelphia, PA, USA}
\affil[9]{School of Clinical Medicine, University of Cambridge,
Cambridge, UK}
\affil[10]{Hospital of the University of Pennsylvania, Philadelphia, PA,
USA}
\affil[11]{Children's Hospital of Philadelphia (CHOP), Philadelphia,
PA, USA}
\affil[12]{Department of Anatomic Pathology and Laboratory Medicine,
Hospital of the University of Pennsylvania, PA, USA}
\affil[13]{Department of Pathology and Laboratory Medicine, University
of California, San Francisco, CA, USA}
\affil[14]{Language Technologies Institute, Carnegie Mellon University,
Pittsburgh, PA, USA}
\affil[15]{Department of Chemistry, Stanford University, Stanford, CA}
\affil[16]{Sport and Exercise Medicine Service, KK Women's and
Children's Hospital, Singapore, Singapore}
\affil[17]{Paediatrics Academic Clinical Programme, SingHealth Duke-NUS
Academic Medical center, Singapore, Singapore}
\affil[18]{College of Computing and Data Science, Nanyang Technological
University, Singapore, Singapore}
\affil[19]{Department of Radiology and Medical Informatics, University
of Geneva, Geneva, Switzerland}
\affil[20]{Department of Biomedical Informatics, Jacobs School of
Medicine, University at Buffalo, Buffalo, NY, United States}
\affil[21]{Department of Biomedical Informatics and Data Science, Yale
School of Medicine, New Haven, CT, USA}
\affil[22]{Division of General Internal Medicine and Primary Care,
Brigham and Women's Hospital, Boston, MA, USA}
\affil[23]{Department of Medicine, Harvard Medical School, Boston, MA,
USA}
\affil[24]{Department of Health Care Policy and Management, Harvard T.
H. Chan School of Public Health, Boston, MA, USA}
\affil[25]{Pre-hospital and Emergency Research Center, Health Services
Research and Population Health, Duke-NUS Medical School, Singapore,
Singapore}
\affil[26]{NUS Artificial Intelligence Institute, National University of
Singapore, Singapore, Singapore}
\affil[27]{Institute of Artificial Intelligence for Digital Health,
Weill Cornell Medicine, New York, NY, USA}
\affil[$\dagger$]{Corresponding authors.}

\begin{document}

\begin{abstract}
Clinical diagnostic evaluation should not only assess whether models
can provide correct diagnoses, but also reflect the realities of
clinical practice, including progressive disclosure of multimodal
information, dynamic updating of diagnostic hypotheses, and continuous
refinement of clinical reasoning. However, existing evaluations of
multimodal large language models (MLLMs) typically rely on single-turn
or isolated tasks, making it difficult to fully capture the complexity
of real-world clinical diagnosis. To bridge this gap, we developed
\clinmm, the largest multi-turn multimodal clinical diagnostic
evaluation benchmark to date. \clinmm contains 1,089 challenging
real-world clinical cases and 3,760 medical images across eight
specialties. We systematically evaluated 15 representative MLLMs using
a two-level evaluation framework that assessed both diagnostic accuracy
and diagnostic reasoning quality. Results showed that proprietary
models achieved the highest overall diagnostic accuracy, but the
proportion of completely correct diagnoses remained limited across all
models. In terms of diagnostic reasoning quality, current models can
identify plausible diagnostic directions but still have considerable
limitations in generating reliable diagnostic reasoning. Error analysis
further identified five representative failure modes: information
synthesis failure, knowledge mapping error, perception error, premature
closure, and visual hallucination.
\end{abstract}


\maketitle

\clearpage
\begingroup
\setlength{\parindent}{0pt}\itshape
\textbf{Corresponding Authors:}\\
Nan Liu, Center for Biomedical Data Science, Duke-NUS Medical School, Singapore, Singapore\\
Email: \href{mailto:liu.nan@duke-nus.edu.sg}{liu.nan@duke-nus.edu.sg}

\medskip
Yifan Peng, Department of Population Health Sciences, Weill Cornell Medicine, New York, NY, USA\\
Email: \href{mailto:yip4002@med.cornell.edu}{yip4002@med.cornell.edu}
\endgroup

\section*{Abstract}
{\AbstractContent}

\section{Introduction}

Diagnostic decision-making lies at the core of clinical practice and
relies on the progressive integration of patient-specific clinical
information from multiple sources~\cite{McDuff2025-om}. In routine
clinical practice, physicians rarely make diagnoses based on a single
symptom, image finding, or lab result. Instead, they update diagnostic
hypotheses as additional information becomes available, reconcile
conflicting evidence, and determine which diagnosis best accounts for
the overall clinical picture~\cite{Scott2009-ud, McMahon2009-qw,
Meyer2013-fc, Centor2019-yp}. Together, these demands make clinical
diagnosis an inherently dynamic and context-dependent
process~\cite{Committee_on_Diagnostic_Error_in_Health_Care2015-uk}.

Recent advances in generative artificial intelligence (AI) have
accelerated its adoption in medicine, with growing potential in clinical
consultation, disease diagnosis, and patient
management~\cite{Yang2023-vj, Fahrner2025-jy, Yang2025-lw, yangRetrievalaugmentedGenerationMedicine}. In
particular, multimodal large language models (MLLMs) have shown
increasing capacity to process multimodal medical data and have achieved
promising performance on tasks such as image interpretation and report
generation~\cite{Tu2024-hd}. As MLLMs continue to improve in domain
knowledge representation and cross-modal understanding, their potential
to support diagnostic decision-making is expected to further
expand~\cite{Saab2026-tb}.

Despite these advancements, it remains unclear whether MLLMs can
perform diagnostic reasoning when clinical information is progressively
disclosed over time and across modalities. Current evaluations are
insufficient to capture this capability~\cite{Johri2025-vn}. First,
existing benchmarks primarily adopt a single-turn, static
question-answering (QA) format, in which complete clinical information
is provided to the model at once to generate an
answer~\cite{Jin2020-bw, Wu2025-ep}. Meanwhile, they typically focus on
isolated tasks, such as image understanding or report
generation~\cite{Zhang2024-ig}. These settings underestimate the
complexity of clinical reasoning because they do not evaluate
progressive multimodal information synthesis, diagnostic hypothesis
revision, cross-turn memory, or the ability to distinguish clinically
grounded reasoning from plausible but hallucinated reasoning. Moreover,
most benchmarks emphasize the accuracy of the diagnoses while
overlooking the quality of the reasoning
process~\cite{McCoy2025-fm}. The reliability of a clinical diagnosis
depends not only on whether the conclusion is correct, but also on
whether the reasoning is grounded in evidence, sufficiently complete,
and logically coherent~\cite{Tanno2025-xs, Omar2025-vp}. Lastly,
although some studies have attempted to construct multi-turn diagnostic
evaluations, they are often limited in scale or not publicly available,
making it difficult to establish a reproducible and extensible evaluation
framework~\cite{McDuff2025-om, Yang2025-ky, nori2025sequential}.

These limitations create a critical gap in the evaluation of MLLMs in
medicine. A model may correctly answer a static medical question yet
fail to synthesize evolving clinical information, misinterpret medical
image findings, hallucinate lab results, or persist with an early
incorrect hypothesis~\cite{Hager2024-cz, Ke2024-aq, Mahajan2025-tr}. On
the other hand, a model may generate a partially correct diagnosis
while omitting essential reasoning steps that are necessary for
clinical trust~\cite{Qiu2025-iv}.

To bridge these gaps, we introduce \clinmm, a multimodal, multi-turn
benchmark designed to evaluate both diagnostic accuracy and reasoning
quality in challenging real-world clinical cases. \clinmm contains
1,089 clinical cases and 3,760 medical images across eight specialties:
Dermatology, Emergency Medicine, Internal Medicine, Nephrology,
Neurology, Oncology, Ophthalmology, and Radiology. Each case is
designed as a multi-turn diagnostic scenario in which clinical
information and images are progressively disclosed, enabling evaluation
of how models synthesize multimodal information and update diagnostic
hypotheses over time.

\begin{figure}[H]
    \centering
    \includegraphics[width=.75\linewidth]{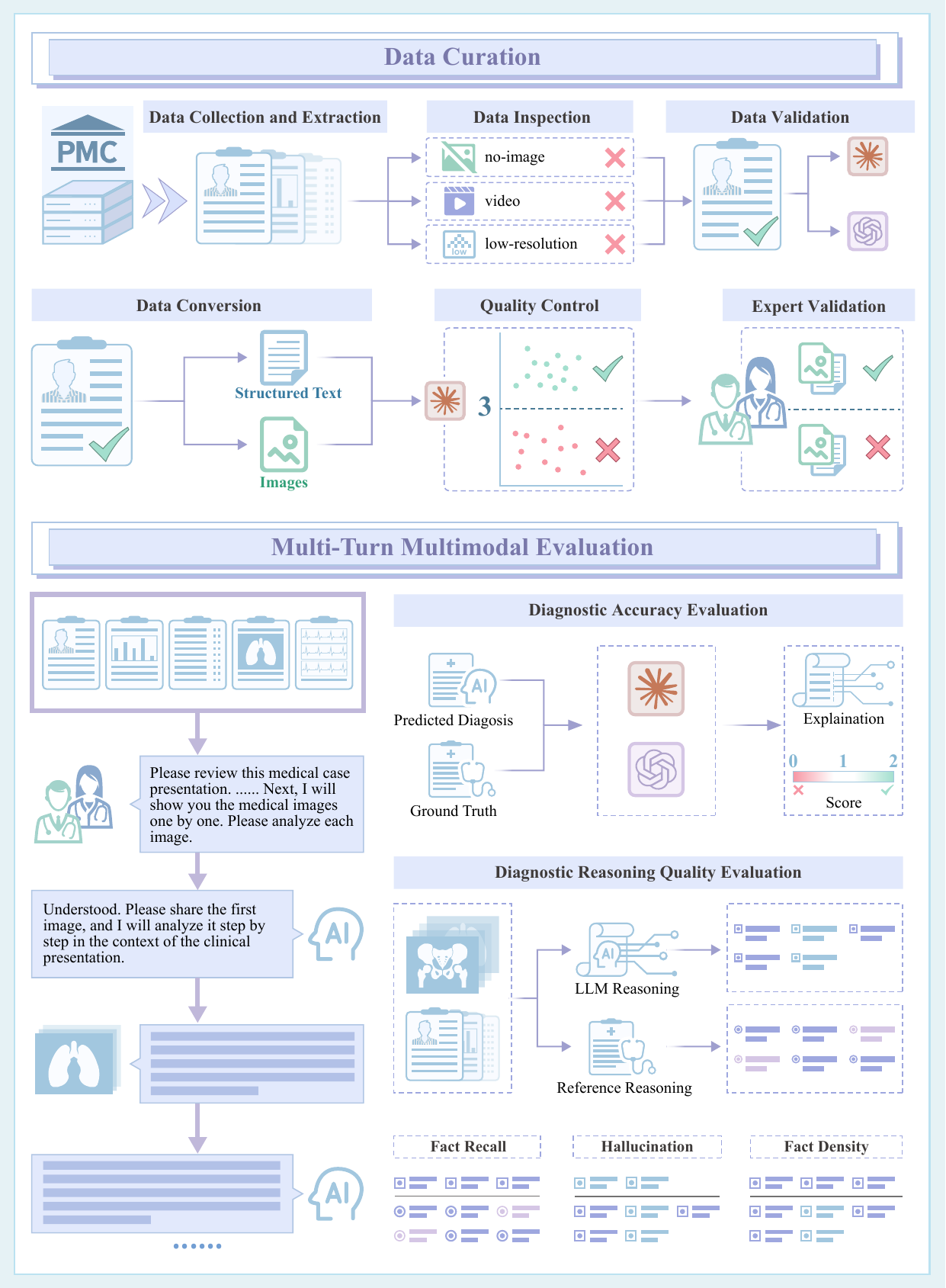}
    \caption{\textbf{Overview of \clinmm and evaluation framework.
    a, Data curation.} \clinmm was developed through a six-stage
    pipeline: (1) Data collection and extraction, in which clinical case
    reports were collected from PubMed Central Open Access; (2) Data
    inspection, where case reports without medical images, those
    containing videos, and those with low-resolution medical images were
    excluded; (3) Data validation, using a dual-LLM consensus mechanism
    to identify cases suitable for the clinical diagnostic reasoning
    task; (4) Data conversion, where validated case reports were parsed
    and transformed into a standardized structured format; (5) Quality
    control, where an automated scoring procedure assessed the
    structured case reports and retained only those meeting a predefined
    quality threshold; and (6) Expert validation, where medical experts
    performed manual verification to ensure data reliability.
    \textbf{b, Multi-turn multimodal evaluation.} During evaluation,
    models perform diagnostic reasoning through multi-turn dialogues,
    with clinical information and images of each case progressively
    disclosed over the course of the dialogue. The evaluation framework
    comprises two levels: (1) Diagnostic accuracy evaluation, in which
    a dual-LLM consensus mechanism compares MLLM-predicted diagnoses
    against the ground-truth diagnoses, with judge LLMs assigning
    accuracy scores ranging from 0 to 2. (2) Diagnostic reasoning
    quality evaluation, where both MLLM-generated reasoning and
    reference reasoning are decomposed into atomic facts and quantified
    across three dimensions: fact recall, hallucination, and fact
    density.}
    \label{fig:overview}
\end{figure}

This study makes three main contributions (\textbf{Fig.~\ref{fig:overview}}):
(1)~We construct \textbf{\clinmm}, to our knowledge, the largest
benchmark to date for multi-turn multimodal diagnostic reasoning on
challenging real-world cases;
(2)~We propose a two-level evaluation framework that measures both
diagnostic accuracy and reasoning quality. Diagnostic accuracy is
assessed using a dual-LLM consensus mechanism, and reasoning quality is
quantified using atomic fact decomposition; fact recall, hallucination,
and fact density measure the completeness, reliability, and efficiency
of MLLMs in clinical diagnostic reasoning, respectively; and
(3)~We systematically evaluate 15 representative MLLMs, covering
proprietary models, open-weight models, medical models, and reasoning
models. Through in-depth analyses, we reveal capability disparities
across different models and identify five representative failure modes
(information synthesis failure, knowledge mapping error, perception
error, premature closure, and visual hallucination), providing important
insights for the future development of MLLMs in medicine.

\section{Results}

\subsection{\clinmm Overview}

\clinmm comprises 1,089 challenging real-world clinical diagnostic
cases curated from PubMed Central Open Access (PMCOA) case reports,
with 3,760 medical images, spanning eight specialties. Radiology is the
largest specialty, with 679 cases (62.35\%), followed by Oncology with
114 cases (10.47\%), Neurology with 98 cases (9.00\%), Dermatology
with 63 cases (5.79\%), Ophthalmology with 60 cases (5.51\%),
Nephrology with 29 cases (2.66\%), Emergency Medicine with 23 cases
(2.11\%), and Internal Medicine with 23 cases (2.11\%). Each case was
structured as a multi-turn multimodal diagnostic dialogue, with an
average of 5.45 dialogue rounds per case. In each case, clinical
information and images were progressively disclosed, requiring models to
update their diagnoses over time. In addition to the ground-truth
diagnosis, each case includes a reference diagnostic reasoning process.
As \clinmm was curated from published case reports, it is enriched for
diagnostically challenging cases involving relatively uncommon clinical
presentations that are underrepresented in routine clinical records.

\subsection{Diagnostic Accuracy}

\subsubsection{Overall Diagnostic Performance of 15 MLLMs on \clinmm}

\begin{figure}[H]
\centering
\includegraphics[width=.8\linewidth]{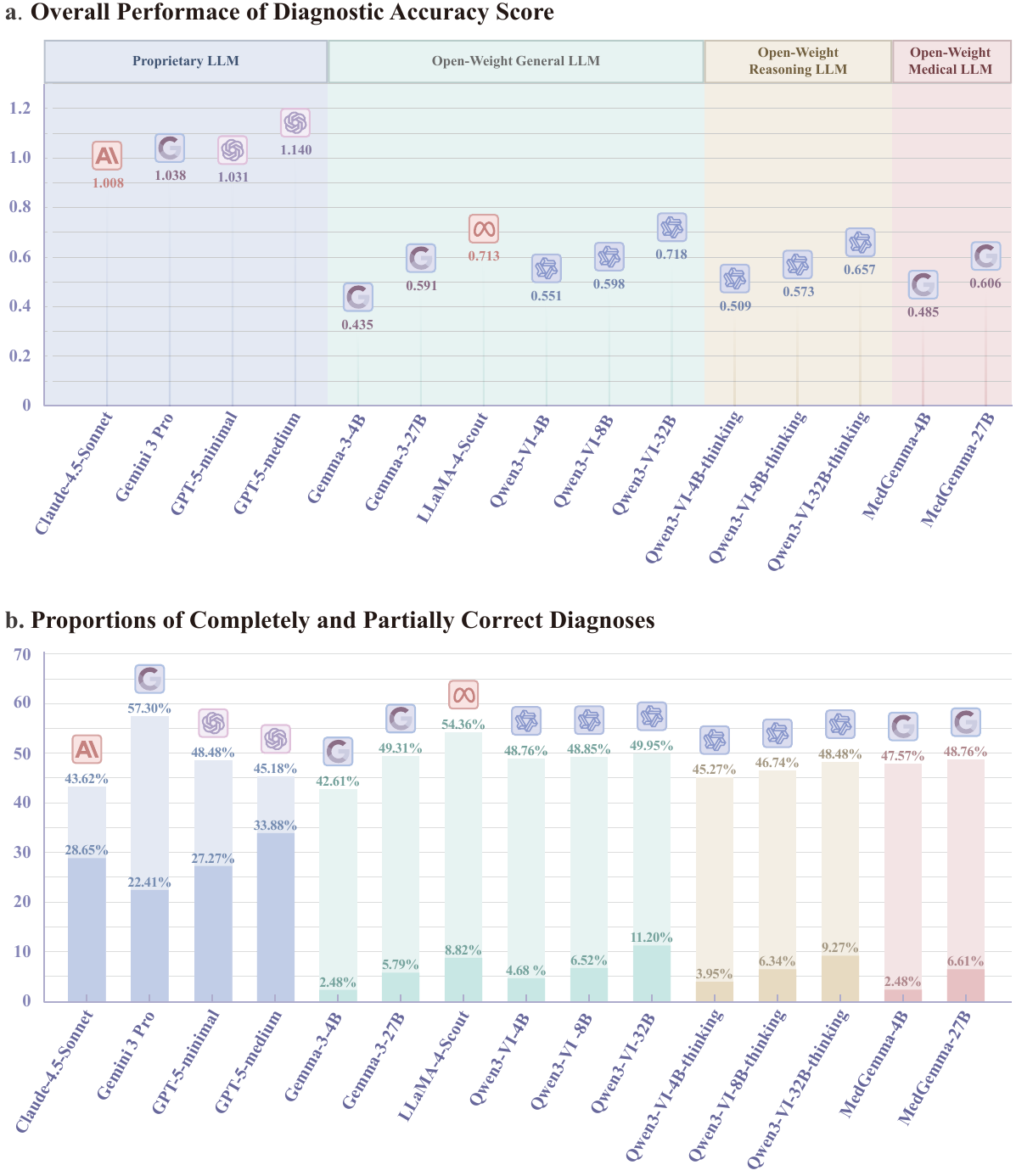}
\caption{\textbf{Overall diagnostic performance of 15 MLLMs on
\clinmm.} \textbf{a, Diagnostic accuracy score.} Mean diagnostic
accuracy scores of 15 representative MLLMs evaluated using the dual-LLM
consensus scoring strategy. Scores range from 0 to 2, with 0 indicating
a completely incorrect or irrelevant diagnosis; 1 indicating a
partially correct diagnosis; and 2 indicating a completely correct
diagnosis. \textbf{b, Proportions of completely and partially correct
diagnoses.} Stacked bar plot showing the proportions of completely
correct and partially correct diagnoses for each model. Darker shading
represents completely correct diagnoses, and lighter shading represents
partially correct diagnoses.}
\label{fig:performance}
\end{figure}

We employed a dual-LLM consensus scoring strategy to evaluate the
overall diagnostic accuracy of 15 representative MLLMs on \clinmm
(\textbf{Fig.~\ref{fig:performance}a}). Scores ranged from 0 to 2, where
0 indicated a completely incorrect or irrelevant diagnosis; 1 indicated
a partially correct diagnosis; and 2 indicated a completely correct
diagnosis. More details are provided in the Methods section under
Diagnostic Accuracy Evaluation.

Proprietary models achieved the strongest overall performance, with
GPT-5-medium obtaining the highest diagnostic accuracy score (1.140),
followed by Gemini 3 Pro (1.038) and GPT-5-minimal (1.031). In contrast,
open-weight models achieved lower performance. Qwen3-VL-32B (0.718) and
LLaMA-4-Scout (0.713) were the two best-performing open-weight models,
but they still showed a gap compared with proprietary models.

We further analyzed the proportions of completely correct (score 2) and
partially correct (0~$<$~consensus score~$<$~2) diagnoses
(\textbf{Fig.~\ref{fig:performance}b}). Among proprietary models,
GPT-5-medium achieved the highest proportion of completely correct
diagnoses (33.88\%), followed by Claude-4.5-Sonnet (28.65\%) and
GPT-5-minimal (27.27\%). Among open-weight models, Qwen3-VL-32B was the
only model with a completely correct diagnosis rate exceeding 10\%
(11.20\%). Meanwhile, across all models, the proportion of partially
correct diagnoses was considerably higher than that of completely
correct diagnoses, a pattern that was particularly pronounced among
open-weight models. For instance, LLaMA-4-Scout achieved a proportion of
partially correct diagnoses of 54.36\%, but only 8.82\% of completely
correct diagnoses. These results indicate that current models often
identify plausible disease categories or related diagnostic directions
but fail to produce precise diagnoses.

Among open-weight models, model scale was positively associated with
improved performance. For the Gemma series, increasing the model size
from 4B to 27B improved the diagnostic accuracy score from 0.435 to
0.591, while the proportion of completely correct diagnoses increased
from 2.48\% to 5.79\%. The MedGemma series showed a similar trend, with
the diagnostic accuracy score increasing from 0.485 to 0.606 and the
proportion of completely correct diagnoses increasing from 2.48\% to
6.61\% as the scale increased from 4B to 27B. In the Qwen3-VL
non-reasoning series, as model size increased from 4B to 8B to 32B,
diagnostic accuracy scores increased from 0.551 at 4B and 0.598 at 8B to
0.718 at 32B, while the proportion of completely correct diagnoses
increased from 4.68\% and 6.52\% to 11.20\%. Similarly, in the Qwen3-VL
reasoning series, diagnostic accuracy scores increased from 0.509 at 4B
and 0.573 at 8B to 0.657 at 32B, and the proportion of completely
correct diagnoses increased from 3.95\% and 6.34\% to 9.27\%. These
results suggest that model scale remains important for multi-turn
multimodal clinical diagnostic reasoning.

\subsubsection{Variations in Diagnostic Accuracy Scores across Specialties}

\begin{figure}[H]
\centering
\includegraphics[width=.9\linewidth]{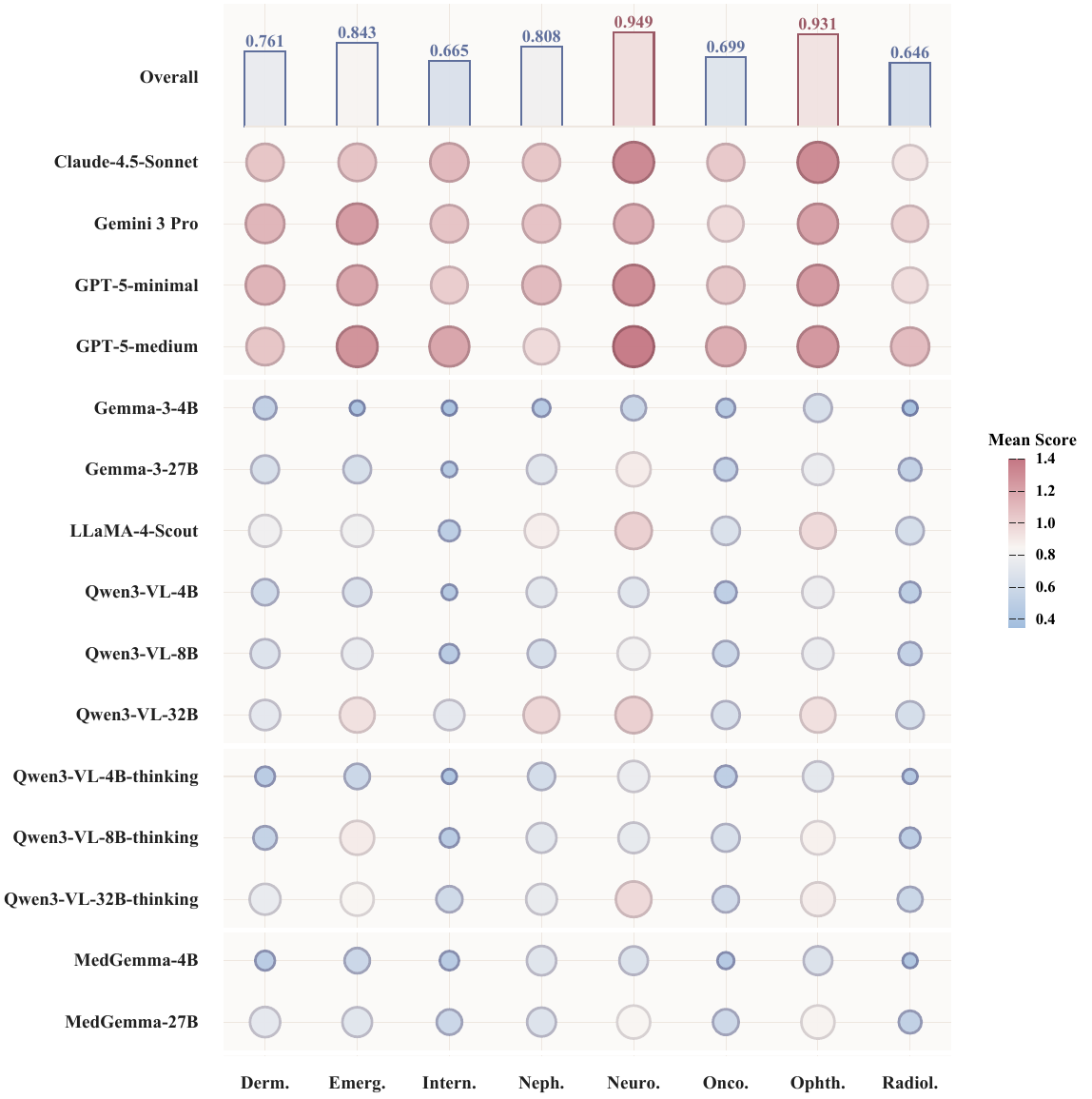}
\caption{\textbf{Diagnostic accuracy scores across specialties on
\clinmm.} Values indicate mean diagnostic accuracy scores, with 95\%
confidence intervals estimated using 100,000 bootstrap resamples. In the
bubble plot, both circle size and color encode the mean diagnostic
accuracy score, with larger and warmer-colored circles indicating higher
scores.}
\label{fig:diagnostic}
\end{figure}

Model performance varied dramatically across specialties
(\textbf{Fig.~\ref{fig:diagnostic}}). Neurology achieved the highest
overall diagnostic accuracy score (0.949 [95\% confidence interval (CI):
0.880, 1.018]), followed by Ophthalmology (0.931 [95\% CI: 0.797,
1.067]), Emergency Medicine (0.843 [95\% CI: 0.677, 1.010]), and
Nephrology (0.808 [95\% CI: 0.666, 0.947]). Intermediate performance was
observed in Dermatology (0.761 [95\% CI: 0.639, 0.884]) and Oncology
(0.699 [95\% CI: 0.626, 0.772]), whereas Internal Medicine (0.665 [95\%
CI: 0.490, 0.851]) and Radiology (0.646 [95\% CI: 0.613, 0.680]) were
the most challenging specialties. Further details are provided in
\textbf{Supplementary Information~\ref{ssec:suppl_diagnostic}}.

\subsubsection{Comparison between General and Medical MLLMs}

\begin{figure}[H]
\centering
\includegraphics[width=.9\linewidth]{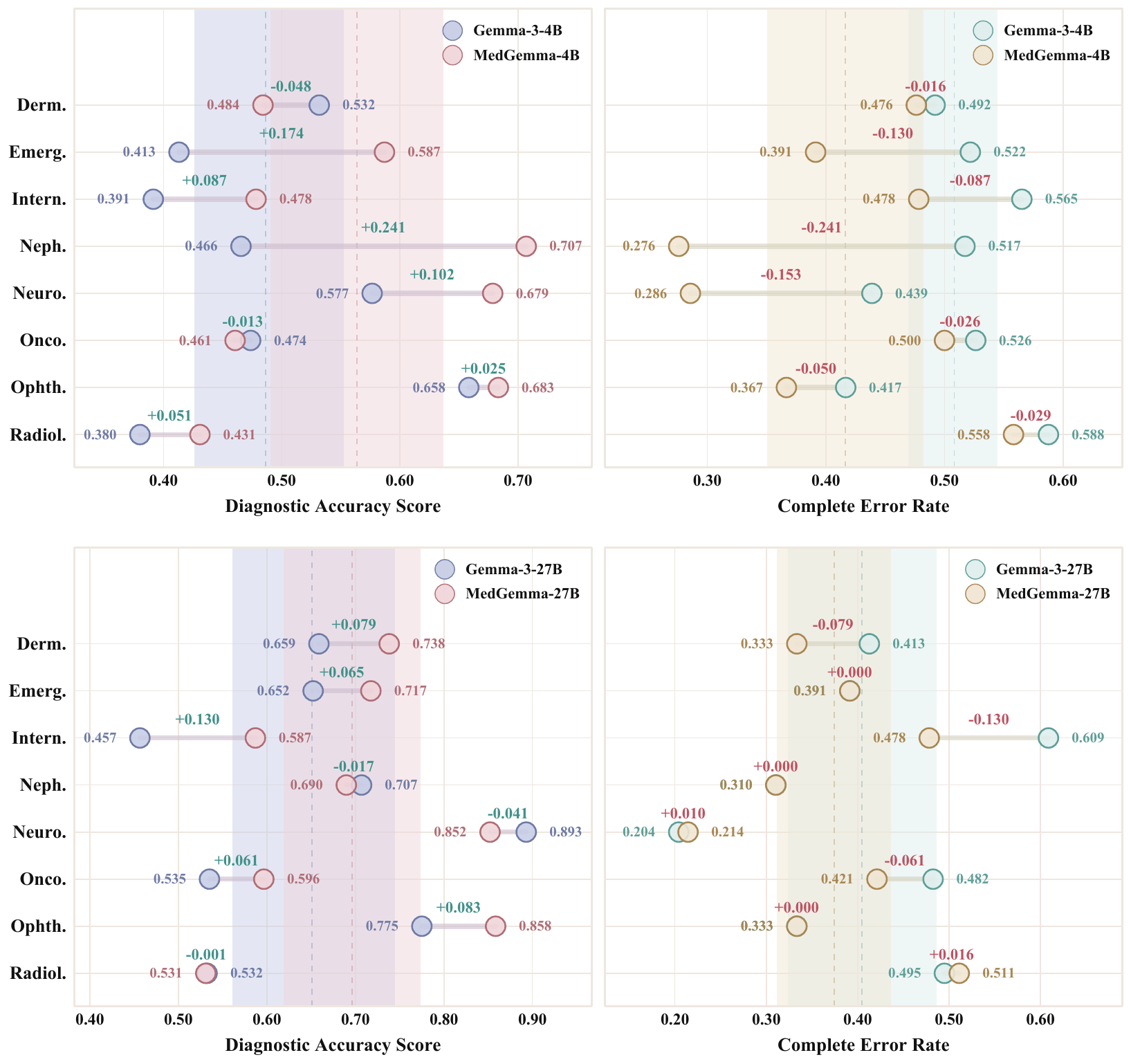}
\caption{\textbf{Comparison between general and medical MLLMs across
specialties on \clinmm.} Diagnostic accuracy scores and complete error
rates of Gemma-3-4B and Gemma-3-27B are compared with their corresponding
medical variants, MedGemma-4B and MedGemma-27B, across eight specialties.
The dashed line indicates the mean value across the eight specialties,
and the shaded band represents the 95\% confidence interval estimated
using 100,000 bootstrap resamples.}
\label{fig:comparison}
\end{figure}

To examine whether medical-domain specialization improves diagnostic
reasoning, we compared Gemma-3-4B and Gemma-3-27B with their corresponding
medical variants, MedGemma-4B and MedGemma-27B
(\textbf{Fig.~\ref{fig:comparison}}). At the 4B scale, MedGemma-4B
outperformed Gemma-3-4B in most specialties, with the largest
improvements in Nephrology ($+$0.241 in overall diagnostic accuracy),
Emergency Medicine ($+$0.174), Neurology ($+$0.102), and Internal
Medicine ($+$0.087). In addition, MedGemma-4B showed lower complete
error rates across all specialties. However, the benefit was less
consistent at the 27B scale. Compared with Gemma-3-27B, MedGemma-27B
improved performance in selected specialties but showed no advantage in
Nephrology, Neurology, and Radiology. These results suggest that
medical-domain adaptation can improve diagnostic performance in smaller
models, but does not uniformly benefit larger models.

\subsubsection{Comparison between Non-Reasoning and Reasoning MLLMs}

\begin{figure}[H]
\centering
\includegraphics[width=.7\linewidth]{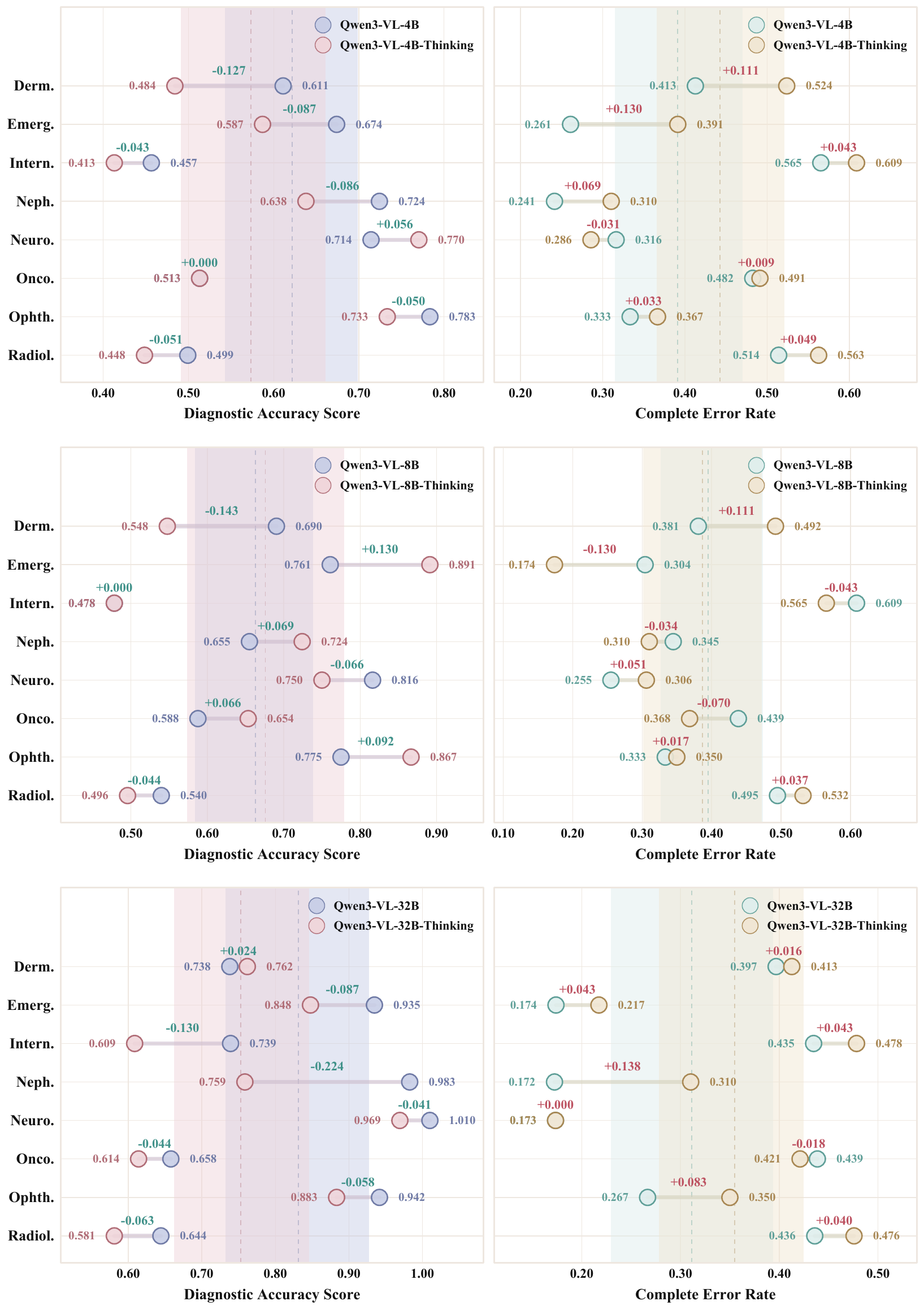}
\caption{\textbf{Comparison between non-reasoning and reasoning MLLMs
across specialties on \clinmm.} Diagnostic accuracy scores and complete
error rates of Qwen3-VL models at the 4B, 8B, and 32B scales under
non-reasoning and reasoning settings across eight specialties. The
dashed line indicates the mean value across the eight specialties, and
the shaded band represents the 95\% confidence interval estimated using
100,000 bootstrap resamples.}
\label{fig:comparison_reasoning}
\end{figure}

To evaluate whether the reasoning setting improves multi-turn multimodal
diagnosis, we compared the non-reasoning and reasoning versions of the
Qwen3-VL series across the 4B, 8B, and 32B scales
(\textbf{Fig.~\ref{fig:comparison_reasoning}}). Overall, the reasoning
setting did not consistently improve diagnostic accuracy. At the 4B
scale, Qwen3-VL-4B-thinking achieved lower diagnostic accuracy scores
than its corresponding non-reasoning variant in most specialties, with
only a small improvement in Neurology ($+$0.056); meanwhile, its
complete error rate was higher in most specialties. At the 8B scale,
the effect of the reasoning setting showed greater specialty-level
heterogeneity, improving diagnostic accuracy scores in Emergency
Medicine, Nephrology, Oncology, and Ophthalmology, but reducing
performance in Dermatology, Neurology, and Radiology. At the 32B scale,
the reasoning variant achieved lower diagnostic accuracy scores than
the non-reasoning variant in most specialties, with the largest decline
observed in Nephrology, where the score decreased from 0.983 to 0.759
($-$0.224); its complete error rate also increased from 0.172 to 0.310
($+$0.138). These findings suggest that simply extending the reasoning
process does not guarantee better outcomes in multi-turn multimodal
clinical reasoning.

\subsection{Diagnostic Reasoning Quality}

\subsubsection{Overall Diagnostic Reasoning Quality of 15 MLLMs on \clinmm}

\begin{figure}[H]
\centering
\includegraphics[width=.9\linewidth]{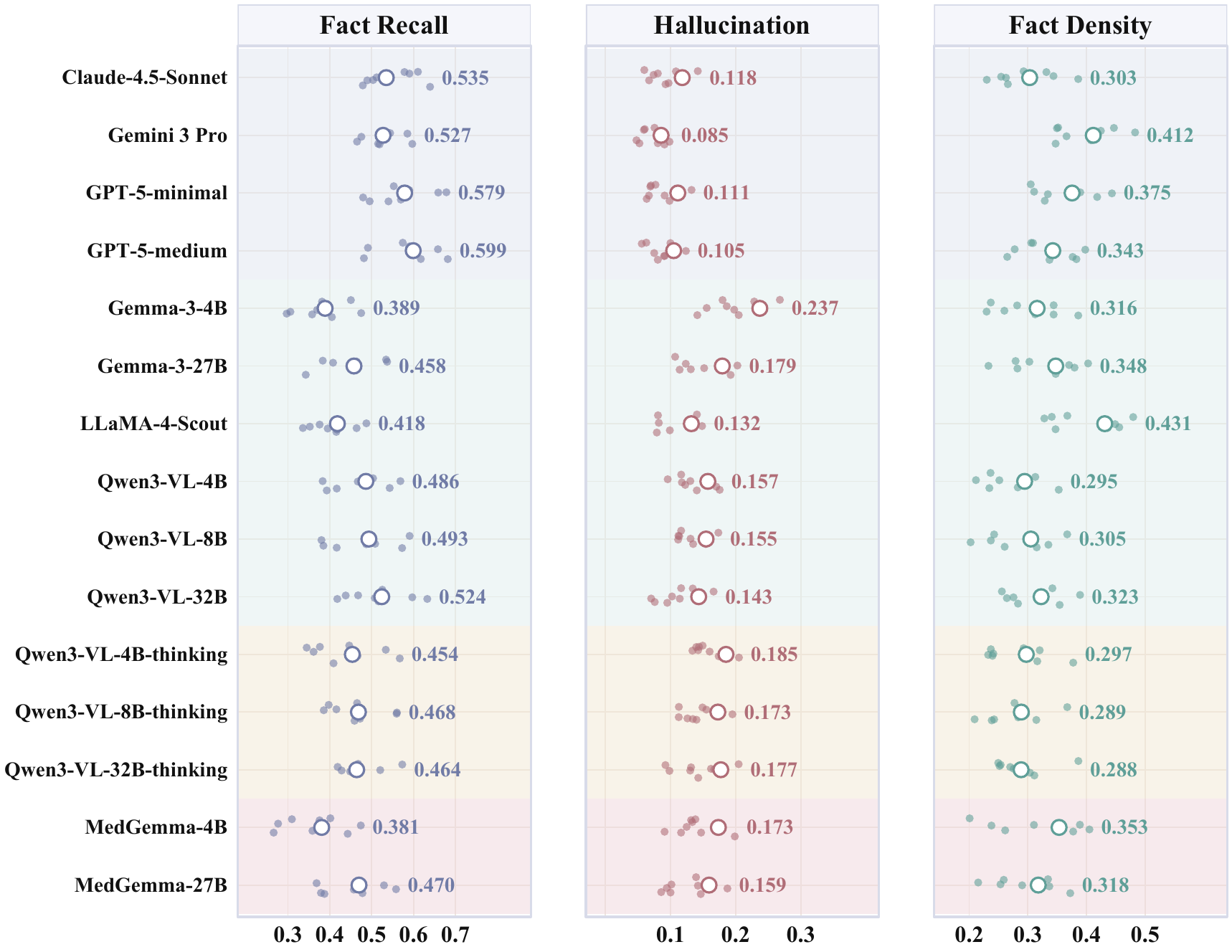}
\caption{\textbf{Overall diagnostic reasoning quality of 15 MLLMs on
\clinmm.} Diagnostic reasoning quality is evaluated through three
metrics: fact recall, hallucination, and fact density. Higher values
indicate better performance for fact recall and fact density, whereas
lower values indicate better performance for hallucination. Open circles
indicate the overall mean values across all cases, and filled dots
indicate the mean values for each specialty.}
\label{fig:diagnostic_reasoning}
\end{figure}

Using atomic fact decomposition, we evaluated the overall quality of
diagnostic reasoning. Model-generated reasoning was decomposed into
atomic clinical facts, enabling assessment of three complementary
dimensions: fact recall, hallucination, and fact density
(\textbf{Fig.~\ref{fig:diagnostic_reasoning}}).

Overall, proprietary models outperformed open-weight models in fact
recall and hallucination control. GPT-5-medium achieved the highest fact
recall (0.599), followed by GPT-5-minimal (0.579), Claude-4.5-Sonnet
(0.535), and Gemini 3 Pro (0.527). For hallucination, Gemini 3 Pro
performed best, achieving the lowest hallucination score (0.085). In
contrast, open-weight models generally showed lower fact recall and
higher hallucination scores. Among open-weight models, Qwen3-VL-32B
achieved the highest fact recall (0.524), whereas LLaMA-4-Scout achieved
the lowest hallucination score (0.132).

For fact density, the results were not fully consistent with fact recall
and hallucination. LLaMA-4-Scout achieved the highest fact density among
all models (0.431). Among proprietary models, Gemini 3 Pro and
GPT-5-minimal showed relatively high fact density, at 0.412 and 0.375,
respectively. Among open-weight models, in addition to LLaMA-4-Scout,
MedGemma-4B (0.353) and Gemma-3-27B (0.348) also achieved relatively
high fact density. Together, these results show that diagnostic
reasoning quality is multidimensional and cannot be captured by a single
score.

\subsubsection{Variations in Fact Recall Scores across Specialties}

\begin{figure}[H]
\centering
\includegraphics[width=\linewidth]{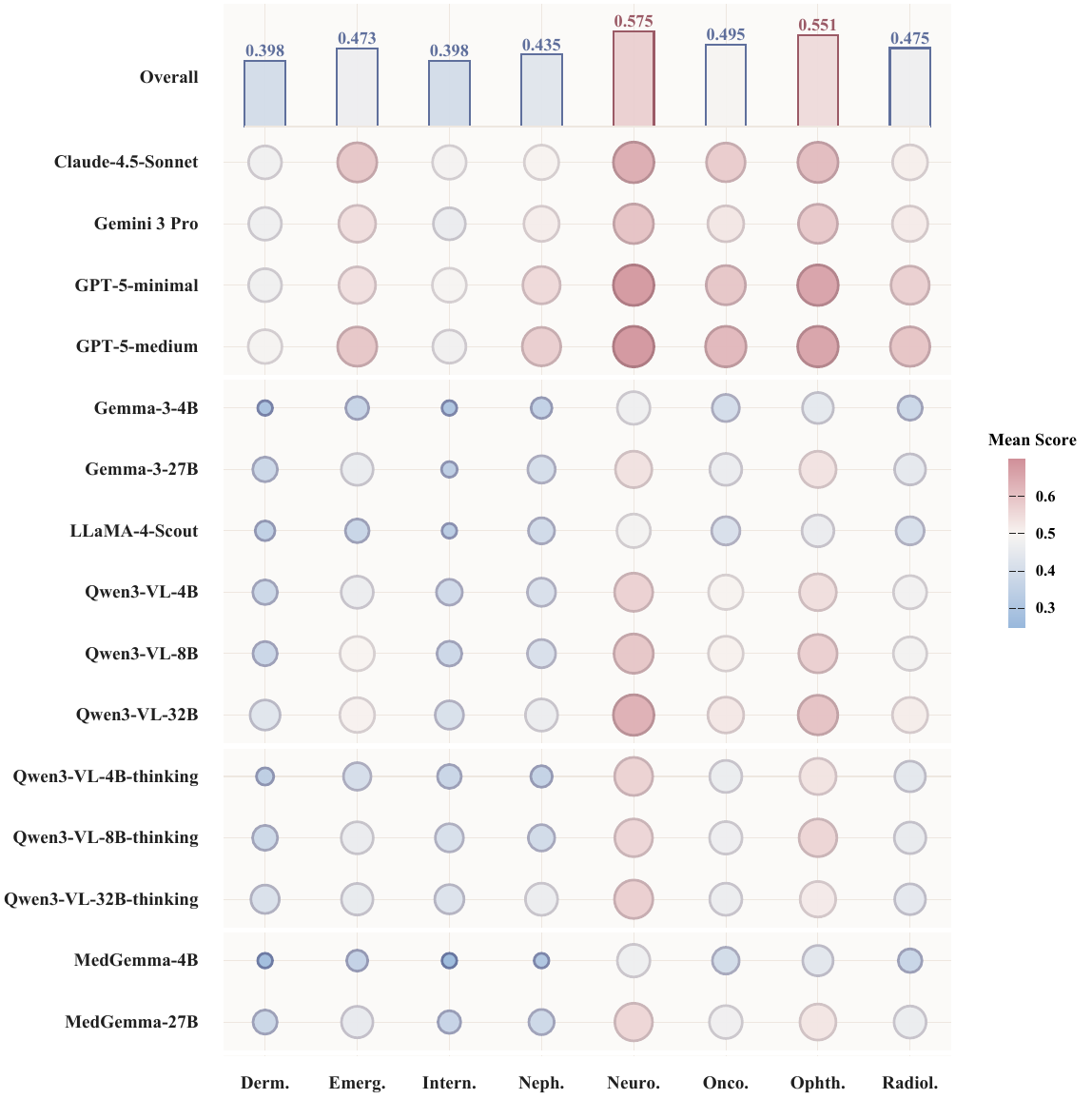}
\caption{\textbf{Fact recall scores across specialties on \clinmm.}
Values indicate mean fact recall scores, with 95\% confidence intervals
estimated using 100,000 bootstrap resamples. In the bubble plot, both
circle size and color encode the mean diagnostic accuracy score, with
larger and warmer-colored circles indicating higher scores.}
\label{fig:fact}
\end{figure}

MLLMs showed notable differences in fact recall scores across
specialties (\textbf{Fig.~\ref{fig:fact}}). Neurology achieved the
highest overall fact recall score (0.575 [95\% CI: 0.550, 0.599]),
followed by Ophthalmology (0.551 [95\% CI: 0.516, 0.584]) and Oncology
(0.495 [95\% CI: 0.470, 0.520]). Intermediate performance was observed
in Radiology (0.475 [95\% CI: 0.465, 0.485]) and Emergency Medicine
(0.473 [95\% CI: 0.417, 0.529]), whereas Nephrology (0.435 [95\% CI:
0.381, 0.489]), Internal Medicine (0.398 [95\% CI: 0.352, 0.444]), and
Dermatology (0.398 [95\% CI: 0.354, 0.441]) showed lower fact recall
scores. These results are broadly consistent with the diagnostic
accuracy analysis and suggest that certain specialties require evidence
types or levels of reasoning complexity that are more difficult for
current models to capture. Further details are provided in
\textbf{Supplementary Information~\ref{ssec:suppl_reasoning}}.

\subsubsection{Effects of Medical Specialization and Reasoning Setting on
Diagnostic Reasoning Quality}

\begin{figure}[H]
\centering
\includegraphics[width=.8\linewidth]{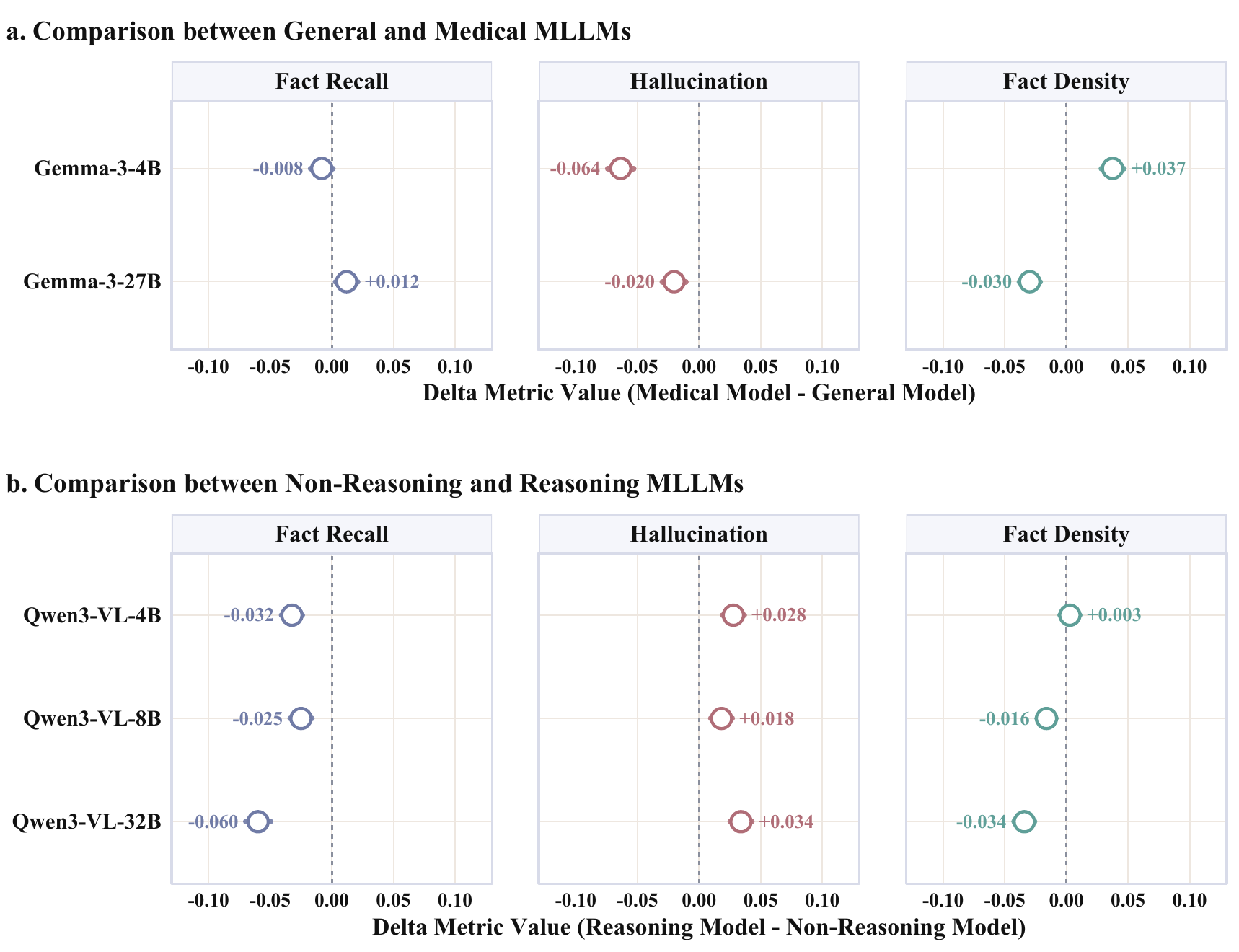}
\caption{\textbf{Effects of medical specialization and reasoning
setting on diagnostic reasoning quality.} \textbf{a, Comparison between
Gemma and MedGemma models.} Delta values of fact recall, hallucination,
and fact density between Gemma models and their corresponding medical
versions at the 4B and 27B scales. \textbf{b, Comparison between
non-reasoning and reasoning MLLMs.} Delta values of fact recall,
hallucination, and fact density between Qwen3-VL models and their
corresponding reasoning versions at the 4B, 8B, and 32B scales. Positive
values indicate better performance for fact recall and fact density,
whereas negative values indicate lower hallucination scores,
corresponding to better performance. Error bars around each point
indicate the 95\% confidence intervals estimated using 100,000 bootstrap
resamples.}
\label{fig:effect}
\end{figure}

We further compared the effects of medical specialization and the
reasoning setting on diagnostic reasoning quality
(\textbf{Fig.~\ref{fig:effect}}). In the comparison between Gemma and
MedGemma, the medical versions had a limited effect on fact recall:
MedGemma-4B showed a slight reduction in fact recall compared with
Gemma-3-4B ($-$0.008), whereas MedGemma-27B showed a small improvement
over Gemma-3-27B ($+$0.012). In contrast, medical specialization reduced
hallucination scores, with delta values of $-$0.064 and $-$0.020 at the
4B and 27B scales, respectively. For fact density, the effects were
inconsistent: MedGemma-4B improved over Gemma-3-4B ($+$0.037), whereas
MedGemma-27B decreased compared with Gemma-3-27B ($-$0.030).

In the Qwen3-VL series, the reasoning setting did not improve
diagnostic reasoning quality. Compared with non-reasoning versions, the
reasoning versions showed lower fact recall at all model scales
($-$0.032, $-$0.025, and $-$0.060) and did not reduce hallucination
scores ($+$0.028, $+$0.018, and $+$0.034). Fact density improved
slightly only for Qwen3-VL-4B-thinking ($+$0.003) and decreased at
larger scales ($-$0.016 and $-$0.034 for 8B and 32B reasoning versions,
respectively).

\subsection{Error Analysis of Multi-Turn Multimodal Clinical Diagnostic
Reasoning}

To investigate the limitations of MLLMs, we analyzed diagnostic failure
cases and categorized them into five representative error types:
information synthesis failure, knowledge mapping error, perception
error, premature closure, and visual hallucination.

First, information synthesis failures occurred when models could not
synthesize information across turns and modalities. For instance, in a
case of smoking-related organizing pneumonia, the model over-relied on
early imaging and cytologic findings suggestive of malignancy,
overlooked subsequent lesion regression and biopsy findings, and
incorrectly diagnosed invasive mucinous adenocarcinoma. Second,
knowledge mapping errors occurred when models recognized key facts but
mapped them to an incorrect disease mechanism. In a case of tickborne
encephalitis, the model recognized fever, cerebrospinal fluid
pleocytosis, MRI abnormalities, and positive intrathecal antibodies but
incorrectly diagnosed acute disseminated encephalomyelitis. Third,
perception errors occurred when models incorrectly interpreted the key
imaging findings. In a case of Chilaiditi syndrome, the model
misinterpreted colonic gas between the liver and diaphragm as gas
within the gallbladder and diagnosed emphysematous cholecystitis.
Fourth, premature closure occurred when models committed too early to
an initial hypothesis and failed to revise it after additional
information emerged. In a case of onychomatricoma, the model continued
to favor subungual melanoma despite subsequent dermoscopic, MRI, and
pathological evidence supporting a benign diagnosis. Finally, visual
hallucinations occurred when models introduced nonexistent imaging
findings. In a case of cerebral cavernoma with hematoma, the model
hallucinated cystic lesions, an eccentric scolex, and a dot sign,
leading to an incorrect diagnosis of neurocysticercosis. Overall, these
errors indicate that diagnostic failures in MLLMs arise from capability
deficits at multiple levels.

\section{Discussion}

\clinmm addresses a critical gap in the evaluation of MLLMs in medicine.
Existing benchmarks largely rely on single-turn or isolated tasks,
whereas clinical diagnosis requires synthesizing information, revising
hypotheses, and reasoning under uncertainty. By converting real-world
case reports into multi-turn, multimodal diagnostic dialogues, \clinmm
enables models to be evaluated in a setting that more closely aligns
with clinical diagnostic workflows. Importantly, the benchmark is
designed to evaluate model performance on diagnostically challenging
cases involving relatively uncommon clinical presentations, rather than
on common diseases seen in routine clinical practice.

Our results show that current MLLMs still have significant limitations
in clinical diagnostic reasoning. Proprietary models outperformed
open-weight models, and most models were able to generate partially
correct diagnoses. However, the proportion of completely correct
diagnoses remained limited; even the best-performing model achieved
completely correct diagnoses in only approximately one-third of cases.
This finding is clinically important: partial recognition may be useful
for triage or differential diagnosis, but reliable diagnostic
decision-making requires accurate synthesis of all relevant
information~\cite{McDuff2025-om,
Committee_on_Diagnostic_Error_in_Health_Care2015-uk}. Specialty-level
differences further indicate that diagnostic performance of MLLMs is
uneven and jointly shaped by the clinical domain, evidence type, and
reasoning complexity~\cite{Zhu2025-fr}.

The analysis of reasoning quality provides additional insight beyond
diagnostic accuracy. Models often omit key evidence, introduce
unsupported statements, or generate inefficient reasoning. These
reasoning issues are particularly relevant for clinical deployment,
where a correct diagnosis without transparent and faithful reasoning
may be difficult to trust, and an incorrect diagnosis supported by
plausible but hallucinated reasoning may be harmful.

In addition, our comparisons reveal important implications for model
development. Larger open-weight models generally achieved higher
diagnostic accuracy and better fact coverage, indicating that scale
remains beneficial for complex multi-turn multimodal diagnosis. Medical
specialization improved some aspects of performance, particularly in
smaller models and in hallucination control, but did not consistently
improve performance in larger models. Notably, the reasoning setting
generally failed to reliably improve diagnostic accuracy or reasoning
quality. This suggests that future progress should not rely solely on
longer reasoning traces~\cite{Hong2025-ot, Kancheti2026-hv}. Instead,
models need better cross-turn memory, stronger visual grounding,
improved clinical knowledge mapping, and mechanisms for revising
hypotheses when additional information contradicts earlier assumptions.

The error analysis further revealed that diagnostic failures do not
arise solely from insufficient medical knowledge, but are distributed
across multiple stages. Perception errors and visual hallucinations
indicate that models' basic image understanding is not yet reliable,
while knowledge mapping errors suggest that even when models identify
key facts, they may fail to accurately connect them to the
corresponding disease mechanisms. Premature closure and information
synthesis failures reflect limitations in models' ability to dynamically
update diagnostic reasoning across multiple turns: models may over-rely
on early clues or initial hypotheses and fail to sufficiently integrate
subsequently disclosed information. These findings highlight the need
for benchmarks that can decompose performance into clinically meaningful
capabilities. \clinmm provides such a framework and can support future
work on model training, evaluation, and safety monitoring.

This study has several limitations. First, \clinmm primarily consists of
challenging published case reports, meaning that its diagnostic
difficulty may differ from that of more common cases encountered in
routine clinical practice~\cite{Vandenbroucke2001-et}. Therefore, this
benchmark should not be taken as a direct estimate of performance on
common clinical diagnoses. Second, \clinmm adopts a passive multi-turn
information disclosure format, whereas in real clinical settings,
physicians actively elicit history, select further investigations, and
dynamically determine the next steps based on existing
hypotheses~\cite{Johri2025-vn}. Third, although we employed multi-stage
data curation, quality control, expert validation, dual-LLM consensus
evaluation, and atomic fact decomposition, potential biases may still
be introduced during data curation and automated evaluation.

Overall, \clinmm establishes a large-scale, multi-turn, multimodal
benchmark for evaluating clinical diagnostic reasoning. Our findings
show that current models can recognize partial diagnostic patterns but
remain unable to reliably achieve completely correct diagnoses or
maintain reasoning fidelity.

\section{Methods}

\subsection{Data Curation}

\clinmm was constructed from clinical case reports published in PMCOA
before September 1, 2025. The search strategy is provided in
\textbf{Supplementary Information~\ref{ssec:search}}. We focused on eight specialties:
Dermatology, Emergency Medicine, Internal Medicine, Nephrology,
Neurology, Oncology, Ophthalmology, and Radiology. The data curation
pipeline consisted of six stages, including Data Collection and
Extraction, Data Inspection, Data Validation, Data Conversion, Data
Quality Control, and Expert Validation. Detailed stage-wise flow
information is provided in \textbf{Supplementary
Information~\ref{ssec:suppl_dataflow}}.

\subsubsection{Data Collection and Extraction}

We established a one-to-one mapping between the manually curated PMID
list and PMCOA, enabling the batch downloading and collection of all
original case reports. On this basis, we performed extraction to obtain
the XML-formatted text and corresponding medical imaging resources.

\subsubsection{Data Inspection}

We excluded case reports that contained videos, lacked medical images,
or had low-resolution medical images below $336\times336$ pixels. This
ensured that retained case reports contained sufficient image
information for MLLMs to perform multimodal diagnostic reasoning.

\subsubsection{Data Validation}

We employed a dual-LLM consensus mechanism with GPT-4.1 and
Claude-4.0-Sonnet to determine whether each case report was suitable for
the diagnostic reasoning task. A valid case report was required to
describe a complete diagnostic process for a single patient, including
essential clinical information, diagnostic medical images, a clearly
defined final diagnosis, and a traceable diagnostic reasoning process.
A case report was retained only when both LLMs classified it as
diagnostically suitable. Case reports focusing solely on treatment
effects, drug reactions, literature reviews, or those lacking
diagnostic reasoning processes were excluded. The detailed prompt is
provided in \textbf{Supplementary Information E}.

\subsubsection{Data Conversion}

We parsed the XML texts of validated case reports and converted them
into a standardized JSON format using GPT-4.1. Extracted key
information included demographic information, chief complaint, history
of present illness, lab results, physical examination, medical images,
diagnosis, and reasoning process. To avoid information leakage,
diagnoses and related findings were removed from the case presentation.
Therefore, models were required to derive the diagnosis through
step-by-step analysis of multi-source information. The detailed prompt
is provided in \textbf{Supplementary Information F}.

\subsubsection{Quality Control}

We used Claude-4.0-Sonnet to compare each structured case against its
original XML text and assigned a fidelity score on a 1--5 scale (5 = all
information perfectly extracted and preserved; 4 = minor discrepancies
without affecting meaning; 3 = some information loss with key elements
preserved; 2 = significant information loss or errors; 1 = critical
information loss or errors). Only cases scoring $\geq$3 were retained,
thereby preventing substantial loss of critical diagnostic
information. The detailed prompt is provided in \textbf{Supplementary
Information G}.

\subsubsection{Expert Validation}

Experienced medical experts (M.W., N.L. (Nicolás Lescano), N.P., E.P.)
reviewed a random sample of the retained cases to verify their clinical
authenticity, the completeness of diagnostic reasoning, and the
effectiveness of information isolation, thereby further ensuring that
\clinmm aligns with the requirements of diagnostic reasoning evaluation.

\subsection{MLLM Evaluation}

We selected a diverse set of 15 representative MLLMs to evaluate their
performance on \clinmm, including proprietary general models, open-weight
general models, open-weight medical models, and reasoning and
non-reasoning variants. Proprietary models included Claude-4.5-Sonnet,
Gemini 3 Pro, and GPT-5, with the latter evaluated under both the
``minimal'' and ``medium'' levels of the ``reasoning effort'' setting.
Open-weight general models comprised the Gemma series (Gemma-3-4B/27B),
LLaMA-4-Scout, and the Qwen3-VL series (Qwen3-VL-4B/8B/32B and their
corresponding reasoning variants). Open-weight medical models included
the MedGemma series (MedGemma-4B/27B).

\subsection{Two-Level Evaluation Framework}

We designed a two-level evaluation framework. The first level assesses
diagnostic accuracy, while the second level evaluates the quality of
the diagnostic reasoning process.

\subsubsection{Diagnostic Accuracy Evaluation}

For diagnostic accuracy evaluation, we adopted a dual-LLM judgment
strategy, in which two LLMs were used as independent evaluators.
Specifically, for each case, we provided both the MLLM-predicted
diagnosis and the ground-truth diagnosis to the judge LLMs (i.e.,
GPT-5-medium and Claude-4.5-Sonnet) and asked each judge LLM to assess
accuracy based on established clinical diagnostic criteria. The scoring
rubric consisted of three levels: 0 for a completely incorrect or
irrelevant diagnosis; 1 for a partially correct diagnosis; and 2 for a
completely correct diagnosis. Additionally, the judge LLMs were
required to provide explicit reasoning to justify their assigned scores.
We then calculated the average of the scores assigned by the two judge
LLMs to obtain a consensus score, thereby reducing single-LLM judgment
bias and allowing the recognition of clinically equivalent diagnostic
expressions with different terminology. The detailed prompt is provided
in \textbf{Supplementary Information H}.

\subsubsection{Diagnostic Reasoning Quality Evaluation}

The quality of reasoning was evaluated using atomic fact decomposition.
Both MLLM-generated reasoning and reference reasoning were decomposed
into a series of atomic clinical facts, with each fact representing a
minimal, independently verifiable clinical statement. We then defined
three metrics: (1) Fact Recall measures reasoning completeness and is
calculated as the proportion of reference atomic facts correctly
identified by the model among all reference atomic facts. A higher fact
recall indicates that the model successfully captures the critical
clinical information required for diagnosis; (2) Hallucination measures
reasoning reliability and is defined as the proportion of model-generated
atomic facts that are unsupported by the reference. A lower
hallucination score reflects stronger adherence to actual clinical
evidence and a reduced tendency to fabricate unsupported information;
and (3) Fact Density measures reasoning efficiency and is quantified as
the proportion of valid atomic facts among all atomic facts produced by
the model. A higher fact density indicates that the model conveys
diagnostic reasoning in a concise and clinically informative manner,
thereby mitigating metric dilution from excessively long outputs. The
detailed prompt is provided in \textbf{Supplementary Information I}.

\section*{Data Availability}
\label{data_availability}
This study used publicly available data from
\url{https://pmc.ncbi.nlm.nih.gov/}.

\section*{Code Availability}
\label{code_availability}
The code used for this study is available at
\url{https://github.com/ruiyang-medinfo/ClinMM}.

\section*{Acknowledgements}
\label{acknowledgements}
This work was supported by the Duke-NUS Signature Research Programme
funded by the Ministry of Health, Singapore. Any opinions, findings and
conclusions or recommendations expressed in this material are those of
the author(s) and do not reflect the views of the Ministry of Health.
This study was supported by the U.S. National Institutes of Health
grants R01LM014573 and R01LM014344.

\section*{Author Contributions}
\label{author_contributions}
Conceptualization: R.Y., N.L. (Nan Liu), Y.P.; Methodology: R.Y.; Software: R.Y.; Data Curation: R.Y.; Investigation: R.Y., W.X., Z.B.; Validation: M.W., N.L. (Nicol\'{a}s Lescano), N.P., E.P.; Formal Analysis: R.Y.; Visualization: R.Y.; Writing – Original Draft: R.Y.; Writing – Review \& Editing: All authors; Supervision: N.L. (Nan Liu), Y.P.; Project Administration: N.L. (Nan Liu), Y.P.; Funding Acquisition: N.L. (Nan Liu), Y.P.

\section*{Competing Interests}
The authors declare no competing interests.

\setlength{\bibsep}{3pt plus 0.3ex}
\bibliographystyle{naturemag}
\bibliography{ref}

\newpage

\appendix

\counterwithin{table}{section}
\renewcommand{\tablename}{Supplementary Table}
\captionsetup[table]{name=Supplementary Table}
\renewcommand\thetable{\Alph{section}\arabic{table}}

\newcommand{\supplementarysection}[2]{%
  \clearpage
  \refstepcounter{section}%
  \section*{Supplementary Information \Alph{section}}%
  \label{#2}%
  \subsection*{#1}%
}

\setlist[enumerate]{nosep,leftmargin=1.5em}
\setlist[itemize]{nosep,leftmargin=1em,label=--}

\supplementarysection{Diagnostic Accuracy across Medical Specialties}{ssec:suppl_diagnostic}

\begin{table}[H]
\centering
\makebox[\linewidth][c]{\resizebox{1.075\linewidth}{!}{%
\begin{tblr}{
  colspec={Q[l,wd=4.15cm] *{8}{Q[c]}},
  rowsep=3.4pt,
  row{1}={bg=Stone1,font=\bfseries},
  column{1}={font=\bfseries,cmd=\mbox},
  hlines = {Stone1},
  vlines = {Stone1}
}
Model & \shortstack{Derm.\\(n=63)} & \shortstack{Emerg.\\(n=23)} & \shortstack{Intern.\\(n=23)} & \shortstack{Neph.\\(n=29)} & \shortstack{Neuro.\\(n=98)} & \shortstack{Onco.\\(n=114)} & \shortstack{Ophth.\\(n=60)} & \shortstack{Radiol.\\(n=679)} \\
Claude-4.5-Sonnet & \shortstack{1.056\\ {[0.865, 1.246]}} & \shortstack{1.065\\ {[0.761, 1.370]}} & \shortstack{1.109\\ {[0.783, 1.435]}} & \shortstack{1.052\\ {[0.862, 1.259]}} & \shortstack{1.327\\ {[1.199, 1.454]}} & \shortstack{1.039\\ {[0.899, 1.175]}} & \shortstack{1.317\\ {[1.133, 1.492]}} & \shortstack{0.918\\ {[0.859, 0.977]}} \\
Gemini 3 Pro & \shortstack{1.127\\ {[0.952, 1.302]}} & \shortstack{1.239\\ {[1.022, 1.457]}} & \shortstack{1.065\\ {[0.739, 1.370]}} & \shortstack{1.069\\ {[0.828, 1.310]}} & \shortstack{1.168\\ {[1.051, 1.286]}} & \shortstack{0.965\\ {[0.851, 1.079]}} & \shortstack{1.217\\ {[1.050, 1.383]}} & \shortstack{0.998\\ {[0.946, 1.049]}} \\
GPT-5-minimal & \shortstack{1.135\\ {[0.944, 1.317]}} & \shortstack{1.196\\ {[0.935, 1.457]}} & \shortstack{1.022\\ {[0.717, 1.326]}} & \shortstack{1.103\\ {[0.879, 1.328]}} & \shortstack{1.311\\ {[1.173, 1.444]}} & \shortstack{1.048\\ {[0.917, 1.180]}} & \shortstack{1.250\\ {[1.067, 1.425]}} & \shortstack{0.951\\ {[0.895, 1.007]}} \\
GPT-5-medium & \shortstack{1.056\\ {[0.849, 1.262]}} & \shortstack{1.283\\ {[1.000, 1.543]}} & \shortstack{1.196\\ {[0.870, 1.522]}} & \shortstack{0.966\\ {[0.724, 1.207]}} & \shortstack{1.383\\ {[1.245, 1.515]}} & \shortstack{1.162\\ {[1.022, 1.303]}} & \shortstack{1.258\\ {[1.075, 1.442]}} & \shortstack{1.099\\ {[1.043, 1.155]}} \\
Gemma-3-4B & \shortstack{0.532\\ {[0.389, 0.675]}} & \shortstack{0.413\\ {[0.239, 0.609]}} & \shortstack{0.391\\ {[0.196, 0.587]}} & \shortstack{0.466\\ {[0.293, 0.638]}} & \shortstack{0.577\\ {[0.464, 0.689]}} & \shortstack{0.474\\ {[0.373, 0.575]}} & \shortstack{0.658\\ {[0.492, 0.833]}} & \shortstack{0.380\\ {[0.343, 0.418]}} \\
Gemma-3-27B & \shortstack{0.659\\ {[0.500, 0.817]}} & \shortstack{0.652\\ {[0.413, 0.891]}} & \shortstack{0.457\\ {[0.217, 0.717]}} & \shortstack{0.707\\ {[0.517, 0.879]}} & \shortstack{0.893\\ {[0.781, 1.005]}} & \shortstack{0.535\\ {[0.430, 0.640]}} & \shortstack{0.775\\ {[0.608, 0.942]}} & \shortstack{0.532\\ {[0.487, 0.577]}} \\
LLaMA-4-Scout & \shortstack{0.802\\ {[0.635, 0.976]}} & \shortstack{0.804\\ {[0.587, 1.022]}} & \shortstack{0.500\\ {[0.283, 0.739]}} & \shortstack{0.879\\ {[0.672, 1.086]}} & \shortstack{1.010\\ {[0.898, 1.122]}} & \shortstack{0.671\\ {[0.557, 0.785]}} & \shortstack{0.967\\ {[0.783, 1.150]}} & \shortstack{0.644\\ {[0.597, 0.690]}} \\
Qwen3-VL-4B & \shortstack{0.611\\ {[0.468, 0.762]}} & \shortstack{0.674\\ {[0.478, 0.891]}} & \shortstack{0.457\\ {[0.217, 0.717]}} & \shortstack{0.724\\ {[0.552, 0.897]}} & \shortstack{0.714\\ {[0.602, 0.832]}} & \shortstack{0.513\\ {[0.417, 0.614]}} & \shortstack{0.783\\ {[0.617, 0.958]}} & \shortstack{0.499\\ {[0.456, 0.543]}} \\
Qwen3-VL-8B & \shortstack{0.690\\ {[0.540, 0.849]}} & \shortstack{0.761\\ {[0.522, 1.022]}} & \shortstack{0.478\\ {[0.217, 0.783]}} & \shortstack{0.655\\ {[0.466, 0.845]}} & \shortstack{0.816\\ {[0.704, 0.929]}} & \shortstack{0.588\\ {[0.482, 0.693]}} & \shortstack{0.775\\ {[0.617, 0.942]}} & \shortstack{0.540\\ {[0.494, 0.586]}} \\
Qwen3-VL-32B & \shortstack{0.738\\ {[0.571, 0.913]}} & \shortstack{0.935\\ {[0.696, 1.174]}} & \shortstack{0.739\\ {[0.457, 1.043]}} & \shortstack{0.983\\ {[0.776, 1.190]}} & \shortstack{1.010\\ {[0.893, 1.128]}} & \shortstack{0.658\\ {[0.535, 0.785]}} & \shortstack{0.942\\ {[0.758, 1.125]}} & \shortstack{0.644\\ {[0.596, 0.694]}} \\

Qwen3-VL-4B-thinking & \shortstack{0.484\\ {[0.349, 0.627]}} & \shortstack{0.587\\ {[0.370, 0.783]}} & \shortstack{0.413\\ {[0.196, 0.652]}} & \shortstack{0.638\\ {[0.466, 0.793]}} & \shortstack{0.770\\ {[0.653, 0.888]}} & \shortstack{0.513\\ {[0.412, 0.614]}} & \shortstack{0.733\\ {[0.567, 0.908]}} & \shortstack{0.448\\ {[0.406, 0.491]}} \\
Qwen3-VL-8B-thinking & \shortstack{0.548\\ {[0.405, 0.698]}} & \shortstack{0.891\\ {[0.674, 1.109]}} & \shortstack{0.478\\ {[0.261, 0.739]}} & \shortstack{0.724\\ {[0.534, 0.914]}} & \shortstack{0.750\\ {[0.633, 0.867]}} & \shortstack{0.654\\ {[0.548, 0.763]}} & \shortstack{0.867\\ {[0.683, 1.058]}} & \shortstack{0.496\\ {[0.451, 0.541]}} \\
Qwen3-VL-32B-thinking & \shortstack{0.762\\ {[0.587, 0.944]}} & \shortstack{0.848\\ {[0.609, 1.087]}} & \shortstack{0.609\\ {[0.348, 0.870]}} & \shortstack{0.759\\ {[0.552, 0.966]}} & \shortstack{0.969\\ {[0.852, 1.087]}} & \shortstack{0.614\\ {[0.504, 0.724]}} & \shortstack{0.883\\ {[0.683, 1.083]}} & \shortstack{0.581\\ {[0.533, 0.629]}} \\
MedGemma-4B & \shortstack{0.484\\ {[0.357, 0.611]}} & \shortstack{0.587\\ {[0.391, 0.783]}} & \shortstack{0.478\\ {[0.283, 0.674]}} & \shortstack{0.707\\ {[0.534, 0.862]}} & \shortstack{0.679\\ {[0.582, 0.776]}} & \shortstack{0.461\\ {[0.373, 0.553]}} & \shortstack{0.683\\ {[0.533, 0.842]}} & \shortstack{0.431\\ {[0.391, 0.471]}} \\
MedGemma-27B & \shortstack{0.738\\ {[0.587, 0.897]}} & \shortstack{0.717\\ {[0.457, 0.978]}} & \shortstack{0.587\\ {[0.326, 0.848]}} & \shortstack{0.690\\ {[0.500, 0.879]}} & \shortstack{0.852\\ {[0.740, 0.964]}} & \shortstack{0.596\\ {[0.496, 0.702]}} & \shortstack{0.858\\ {[0.683, 1.033]}} & \shortstack{0.531\\ {[0.485, 0.577]}} \\
Overall & \shortstack{0.761\\ {[0.639, 0.884]}} & \shortstack{0.843\\ {[0.677, 1.010]}} & \shortstack{0.665\\ {[0.490, 0.851]}} & \shortstack{0.808\\ {[0.666, 0.947]}} & \shortstack{0.949\\ {[0.880, 1.018]}} & \shortstack{0.699\\ {[0.626, 0.772]}} & \shortstack{0.931\\ {[0.797, 1.067]}} & \shortstack{0.646\\ {[0.613, 0.680]}} \\
\end{tblr}
}}
\caption{\textbf{Diagnostic accuracy scores across medical specialties
on ClinMM-Bench.} Values indicate mean diagnostic accuracy scores,
with 95\% confidence intervals estimated using 100,000 bootstrap
resamples.}
\label{stab:diagnostic}
\end{table}

\begin{table}[H]
\centering
\makebox[\linewidth][c]{\resizebox{1.075\linewidth}{!}{%
\begin{tblr}{
  colspec={Q[l,wd=4.15cm] *{8}{Q[c]}},
  rowsep=3.4pt,
  row{1}={bg=Stone1,font=\bfseries},
  column{1}={font=\bfseries,cmd=\mbox},
  hlines = {Stone1},
  vlines = {Stone1}
}
Model & \shortstack{Derm.\\(n=63)} & \shortstack{Emerg.\\(n=23)} & \shortstack{Intern.\\(n=23)} & \shortstack{Neph.\\(n=29)} & \shortstack{Neuro.\\(n=98)} & \shortstack{Onco.\\(n=114)} & \shortstack{Ophth.\\(n=60)} & \shortstack{Radiol.\\(n=679)} \\
Claude-4.5-Sonnet & \shortstack{0.317\\ {[0.206, 0.429]}} & \shortstack{0.304\\ {[0.130, 0.478]}} & \shortstack{0.391\\ {[0.217, 0.609]}} & \shortstack{0.172\\ {[0.034, 0.310]}} & \shortstack{0.398\\ {[0.306, 0.500]}} & \shortstack{0.281\\ {[0.202, 0.368]}} & \shortstack{0.433\\ {[0.317, 0.567]}} & \shortstack{0.256\\ {[0.224, 0.290]}} \\

Gemini 3 Pro & \shortstack{0.270\\ {[0.159, 0.381]}} & \shortstack{0.261\\ {[0.087, 0.435]}} & \shortstack{0.304\\ {[0.130, 0.478]}} & \shortstack{0.241\\ {[0.103, 0.414]}} & \shortstack{0.235\\ {[0.153, 0.316]}} & \shortstack{0.167\\ {[0.105, 0.237]}} & \shortstack{0.317\\ {[0.200, 0.433]}} & \shortstack{0.215\\ {[0.184, 0.246]}} \\

GPT-5-minimal & \shortstack{0.333\\ {[0.222, 0.444]}} & \shortstack{0.304\\ {[0.130, 0.478]}} & \shortstack{0.304\\ {[0.130, 0.478]}} & \shortstack{0.241\\ {[0.103, 0.414]}} & \shortstack{0.408\\ {[0.316, 0.510]}} & \shortstack{0.263\\ {[0.184, 0.342]}} & \shortstack{0.367\\ {[0.250, 0.483]}} & \shortstack{0.240\\ {[0.208, 0.272]}} \\

GPT-5-medium & \shortstack{0.349\\ {[0.238, 0.476]}} & \shortstack{0.391\\ {[0.217, 0.609]}} & \shortstack{0.435\\ {[0.217, 0.652]}} & \shortstack{0.207\\ {[0.069, 0.345]}} & \shortstack{0.480\\ {[0.378, 0.582]}} & \shortstack{0.360\\ {[0.272, 0.447]}} & \shortstack{0.400\\ {[0.283, 0.533]}} & \shortstack{0.309\\ {[0.275, 0.345]}} \\

Gemma-3-4B & \shortstack{0.048\\ {[0.000, 0.111]}} & \shortstack{0.000\\ {[0.000, 0.000]}} & \shortstack{0.000\\ {[0.000, 0.000]}} & \shortstack{0.000\\ {[0.000, 0.000]}} & \shortstack{0.051\\ {[0.010, 0.102]}} & \shortstack{0.035\\ {[0.009, 0.070]}} & \shortstack{0.100\\ {[0.033, 0.183]}} & \shortstack{0.013\\ {[0.006, 0.022]}} \\

Gemma-3-27B & \shortstack{0.095\\ {[0.032, 0.175]}} & \shortstack{0.043\\ {[0.000, 0.130]}} & \shortstack{0.043\\ {[0.000, 0.130]}} & \shortstack{0.000\\ {[0.000, 0.000]}} & \shortstack{0.092\\ {[0.041, 0.153]}} & \shortstack{0.035\\ {[0.009, 0.070]}} & \shortstack{0.117\\ {[0.050, 0.200]}} & \shortstack{0.052\\ {[0.035, 0.069]}} \\

LLaMA-4-Scout & \shortstack{0.143\\ {[0.063, 0.238]}} & \shortstack{0.043\\ {[0.000, 0.130]}} & \shortstack{0.043\\ {[0.000, 0.130]}} & \shortstack{0.103\\ {[0.000, 0.207]}} & \shortstack{0.163\\ {[0.092, 0.235]}} & \shortstack{0.061\\ {[0.018, 0.105]}} & \shortstack{0.200\\ {[0.100, 0.300]}} & \shortstack{0.069\\ {[0.050, 0.088]}} \\

Qwen3-VL-4B & \shortstack{0.063\\ {[0.016, 0.127]}} & \shortstack{0.043\\ {[0.000, 0.130]}} & \shortstack{0.043\\ {[0.000, 0.130]}} & \shortstack{0.034\\ {[0.000, 0.103]}} & \shortstack{0.061\\ {[0.020, 0.112]}} & \shortstack{0.018\\ {[0.000, 0.044]}} & \shortstack{0.100\\ {[0.033, 0.183]}} & \shortstack{0.044\\ {[0.029, 0.060]}} \\

Qwen3-VL-8B & \shortstack{0.079\\ {[0.016, 0.159]}} & \shortstack{0.087\\ {[0.000, 0.217]}} & \shortstack{0.130\\ {[0.000, 0.261]}} & \shortstack{0.034\\ {[0.000, 0.103]}} & \shortstack{0.061\\ {[0.020, 0.112]}} & \shortstack{0.044\\ {[0.009, 0.088]}} & \shortstack{0.100\\ {[0.033, 0.183]}} & \shortstack{0.063\\ {[0.046, 0.082]}} \\

Qwen3-VL-32B & \shortstack{0.127\\ {[0.048, 0.206]}} & \shortstack{0.130\\ {[0.000, 0.261]}} & \shortstack{0.130\\ {[0.000, 0.261]}} & \shortstack{0.138\\ {[0.034, 0.276]}} & \shortstack{0.163\\ {[0.092, 0.235]}} & \shortstack{0.114\\ {[0.061, 0.175]}} & \shortstack{0.233\\ {[0.133, 0.350]}} & \shortstack{0.090\\ {[0.069, 0.112]}} \\

Qwen3-VL-4B-thinking & \shortstack{0.032\\ {[0.000, 0.079]}} & \shortstack{0.000\\ {[0.000, 0.000]}} & \shortstack{0.000\\ {[0.000, 0.000]}} & \shortstack{0.000\\ {[0.000, 0.000]}} & \shortstack{0.082\\ {[0.031, 0.143]}} & \shortstack{0.026\\ {[0.000, 0.061]}} & \shortstack{0.117\\ {[0.050, 0.200]}} & \shortstack{0.034\\ {[0.021, 0.049]}} \\

Qwen3-VL-8B-thinking & \shortstack{0.063\\ {[0.016, 0.127]}} & \shortstack{0.087\\ {[0.000, 0.217]}} & \shortstack{0.043\\ {[0.000, 0.130]}} & \shortstack{0.034\\ {[0.000, 0.103]}} & \shortstack{0.071\\ {[0.020, 0.122]}} & \shortstack{0.061\\ {[0.018, 0.105]}} & \shortstack{0.183\\ {[0.083, 0.283]}} & \shortstack{0.053\\ {[0.037, 0.071]}} \\

Qwen3-VL-32B-thinking & \shortstack{0.159\\ {[0.079, 0.254]}} & \shortstack{0.130\\ {[0.000, 0.261]}} & \shortstack{0.087\\ {[0.000, 0.217]}} & \shortstack{0.069\\ {[0.000, 0.172]}} & \shortstack{0.143\\ {[0.082, 0.214]}} & \shortstack{0.053\\ {[0.018, 0.096]}} & \shortstack{0.250\\ {[0.150, 0.367]}} & \shortstack{0.072\\ {[0.053, 0.093]}} \\

MedGemma-4B & \shortstack{0.016\\ {[0.000, 0.048]}} & \shortstack{0.000\\ {[0.000, 0.000]}} & \shortstack{0.000\\ {[0.000, 0.000]}} & \shortstack{0.000\\ {[0.000, 0.000]}} & \shortstack{0.010\\ {[0.000, 0.031]}} & \shortstack{0.009\\ {[0.000, 0.026]}} & \shortstack{0.083\\ {[0.017, 0.167]}} & \shortstack{0.028\\ {[0.016, 0.041]}} \\

MedGemma-27B & \shortstack{0.111\\ {[0.048, 0.190]}} & \shortstack{0.087\\ {[0.000, 0.217]}} & \shortstack{0.087\\ {[0.000, 0.217]}} & \shortstack{0.034\\ {[0.000, 0.103]}} & \shortstack{0.082\\ {[0.031, 0.143]}} & \shortstack{0.035\\ {[0.009, 0.070]}} & \shortstack{0.133\\ {[0.050, 0.217]}} & \shortstack{0.059\\ {[0.041, 0.077]}} \\

Overall & \shortstack{0.147\\ {[0.098, 0.200]}} & \shortstack{0.128\\ {[0.064, 0.203]}} & \shortstack{0.136\\ {[0.070, 0.212]}} & \shortstack{0.087\\ {[0.037, 0.149]}} & \shortstack{0.167\\ {[0.131, 0.205]}} & \shortstack{0.104\\ {[0.080, 0.130]}} & \shortstack{0.209\\ {[0.143, 0.280]}} & \shortstack{0.107\\ {[0.094, 0.120]}} \\
\end{tblr}
}}
\caption{\textbf{Proportions of completely correct diagnoses across
medical specialties on ClinMM-Bench.} Values indicate proportions of
completely correct diagnoses, with 95\% confidence intervals estimated
using 100,000 bootstrap resamples.}
\label{stab:proportion complete}
\end{table}

\begin{table}[H]
\centering
\makebox[\linewidth][c]{\resizebox{1.075\linewidth}{!}{%
\begin{tblr}{
  colspec={Q[l,wd=4.15cm] *{8}{Q[c]}},
  rowsep=3.4pt,
  row{1}={bg=Stone1,font=\bfseries},
  column{1}={font=\bfseries,cmd=\mbox},
  hlines = {Stone1},
  vlines = {Stone1}
}
Model & \shortstack{Derm.\\(n=63)} & \shortstack{Emerg.\\(n=23)} & \shortstack{Intern.\\(n=23)} & \shortstack{Neph.\\(n=29)} & \shortstack{Neuro.\\(n=98)} & \shortstack{Onco.\\(n=114)} & \shortstack{Ophth.\\(n=60)} & \shortstack{Radiol.\\(n=679)} \\

Claude-4.5-Sonnet & \shortstack{0.413\\ {[0.286, 0.540]}} & \shortstack{0.478\\ {[0.261, 0.696]}} & \shortstack{0.348\\ {[0.174, 0.565]}} & \shortstack{0.759\\ {[0.586, 0.897]}} & \shortstack{0.520\\ {[0.418, 0.622]}} & \shortstack{0.465\\ {[0.377, 0.553]}} & \shortstack{0.450\\ {[0.333, 0.583]}} & \shortstack{0.408\\ {[0.371, 0.445]}} \\

Gemini 3 Pro & \shortstack{0.524\\ {[0.397, 0.651]}} & \shortstack{0.696\\ {[0.522, 0.870]}} & \shortstack{0.435\\ {[0.217, 0.652]}} & \shortstack{0.552\\ {[0.379, 0.724]}} & \shortstack{0.653\\ {[0.561, 0.745]}} & \shortstack{0.632\\ {[0.544, 0.719]}} & \shortstack{0.550\\ {[0.417, 0.667]}} & \shortstack{0.560\\ {[0.523, 0.596]}} \\

GPT-5-minimal & \shortstack{0.444\\ {[0.317, 0.571]}} & \shortstack{0.609\\ {[0.391, 0.783]}} & \shortstack{0.435\\ {[0.217, 0.652]}} & \shortstack{0.621\\ {[0.448, 0.793]}} & \shortstack{0.469\\ {[0.367, 0.571]}} & \shortstack{0.518\\ {[0.430, 0.605]}} & \shortstack{0.467\\ {[0.333, 0.600]}} & \shortstack{0.479\\ {[0.440, 0.517]}} \\

GPT-5-medium & \shortstack{0.333\\ {[0.222, 0.460]}} & \shortstack{0.522\\ {[0.304, 0.739]}} & \shortstack{0.348\\ {[0.174, 0.565]}} & \shortstack{0.552\\ {[0.379, 0.724]}} & \shortstack{0.418\\ {[0.327, 0.520]}} & \shortstack{0.430\\ {[0.342, 0.518]}} & \shortstack{0.450\\ {[0.333, 0.583]}} & \shortstack{0.468\\ {[0.432, 0.507]}} \\

Gemma-3-4B & \shortstack{0.460\\ {[0.333, 0.587]}} & \shortstack{0.478\\ {[0.261, 0.696]}} & \shortstack{0.435\\ {[0.217, 0.652]}} & \shortstack{0.483\\ {[0.310, 0.655]}} & \shortstack{0.510\\ {[0.408, 0.612]}} & \shortstack{0.439\\ {[0.351, 0.526]}} & \shortstack{0.483\\ {[0.367, 0.617]}} & \shortstack{0.399\\ {[0.362, 0.436]}} \\

Gemma-3-27B & \shortstack{0.492\\ {[0.365, 0.619]}} & \shortstack{0.565\\ {[0.348, 0.783]}} & \shortstack{0.348\\ {[0.174, 0.565]}} & \shortstack{0.690\\ {[0.517, 0.862]}} & \shortstack{0.704\\ {[0.612, 0.796]}} & \shortstack{0.482\\ {[0.395, 0.570]}} & \shortstack{0.550\\ {[0.417, 0.667]}} & \shortstack{0.454\\ {[0.417, 0.492]}} \\

LLaMA-4-Scout & \shortstack{0.524\\ {[0.397, 0.651]}} & \shortstack{0.696\\ {[0.522, 0.870]}} & \shortstack{0.435\\ {[0.217, 0.652]}} & \shortstack{0.690\\ {[0.517, 0.862]}} & \shortstack{0.704\\ {[0.612, 0.796]}} & \shortstack{0.535\\ {[0.439, 0.623]}} & \shortstack{0.517\\ {[0.383, 0.650]}} & \shortstack{0.518\\ {[0.482, 0.555]}} \\

Qwen3-VL-4B & \shortstack{0.524\\ {[0.397, 0.651]}} & \shortstack{0.696\\ {[0.522, 0.870]}} & \shortstack{0.391\\ {[0.217, 0.609]}} & \shortstack{0.724\\ {[0.552, 0.862]}} & \shortstack{0.622\\ {[0.531, 0.714]}} & \shortstack{0.500\\ {[0.412, 0.588]}} & \shortstack{0.567\\ {[0.433, 0.683]}} & \shortstack{0.442\\ {[0.405, 0.479]}} \\

Qwen3-VL-8B & \shortstack{0.540\\ {[0.413, 0.667]}} & \shortstack{0.609\\ {[0.391, 0.783]}} & \shortstack{0.261\\ {[0.087, 0.435]}} & \shortstack{0.621\\ {[0.448, 0.793]}} & \shortstack{0.684\\ {[0.592, 0.776]}} & \shortstack{0.518\\ {[0.430, 0.605]}} & \shortstack{0.567\\ {[0.433, 0.683]}} & \shortstack{0.442\\ {[0.405, 0.479]}} \\

Qwen3-VL-32B & \shortstack{0.476\\ {[0.349, 0.603]}} & \shortstack{0.696\\ {[0.522, 0.870]}} & \shortstack{0.435\\ {[0.217, 0.652]}} & \shortstack{0.690\\ {[0.517, 0.862]}} & \shortstack{0.663\\ {[0.571, 0.755]}} & \shortstack{0.447\\ {[0.360, 0.535]}} & \shortstack{0.500\\ {[0.367, 0.633]}} & \shortstack{0.474\\ {[0.437, 0.513]}} \\

Qwen3-VL-4B-thinking & \shortstack{0.444\\ {[0.317, 0.571]}} & \shortstack{0.609\\ {[0.391, 0.783]}} & \shortstack{0.391\\ {[0.217, 0.609]}} & \shortstack{0.690\\ {[0.517, 0.862]}} & \shortstack{0.633\\ {[0.541, 0.724]}} & \shortstack{0.482\\ {[0.395, 0.570]}} & \shortstack{0.517\\ {[0.383, 0.650]}} & \shortstack{0.404\\ {[0.367, 0.440]}} \\

Qwen3-VL-8B-thinking & \shortstack{0.444\\ {[0.317, 0.571]}} & \shortstack{0.739\\ {[0.565, 0.913]}} & \shortstack{0.391\\ {[0.217, 0.609]}} & \shortstack{0.655\\ {[0.483, 0.828]}} & \shortstack{0.622\\ {[0.531, 0.714]}} & \shortstack{0.570\\ {[0.482, 0.658]}} & \shortstack{0.467\\ {[0.350, 0.600]}} & \shortstack{0.415\\ {[0.378, 0.452]}} \\

Qwen3-VL-32B-thinking & \shortstack{0.429\\ {[0.302, 0.556]}} & \shortstack{0.652\\ {[0.435, 0.826]}} & \shortstack{0.435\\ {[0.217, 0.652]}} & \shortstack{0.621\\ {[0.448, 0.793]}} & \shortstack{0.684\\ {[0.592, 0.776]}} & \shortstack{0.526\\ {[0.439, 0.614]}} & \shortstack{0.400\\ {[0.283, 0.517]}} & \shortstack{0.452\\ {[0.415, 0.490]}} \\

MedGemma-4B & \shortstack{0.508\\ {[0.381, 0.635]}} & \shortstack{0.609\\ {[0.391, 0.783]}} & \shortstack{0.522\\ {[0.304, 0.739]}} & \shortstack{0.724\\ {[0.552, 0.862]}} & \shortstack{0.704\\ {[0.612, 0.796]}} & \shortstack{0.491\\ {[0.395, 0.579]}} & \shortstack{0.550\\ {[0.417, 0.667]}} & \shortstack{0.414\\ {[0.377, 0.451]}} \\

MedGemma-27B & \shortstack{0.556\\ {[0.429, 0.683]}} & \shortstack{0.522\\ {[0.304, 0.739]}} & \shortstack{0.435\\ {[0.217, 0.652]}} & \shortstack{0.655\\ {[0.483, 0.828]}} & \shortstack{0.704\\ {[0.612, 0.796]}} & \shortstack{0.544\\ {[0.456, 0.632]}} & \shortstack{0.533\\ {[0.400, 0.667]}} & \shortstack{0.430\\ {[0.393, 0.467]}} \\

Overall & \shortstack{0.474\\ {[0.393, 0.556]}} & \shortstack{0.612\\ {[0.501, 0.719]}} & \shortstack{0.403\\ {[0.287, 0.528]}} & \shortstack{0.648\\ {[0.538, 0.752]}} & \shortstack{0.620\\ {[0.569, 0.670]}} & \shortstack{0.505\\ {[0.449, 0.563]}} & \shortstack{0.504\\ {[0.429, 0.579]}} & \shortstack{0.451\\ {[0.427, 0.475]}} \\
\end{tblr}
}}
\caption{\textbf{Proportions of partially correct diagnoses across
medical specialties on ClinMM-Bench.} Values indicate proportions of
partially correct diagnoses, with 95\% confidence intervals estimated
using 100,000 bootstrap resamples.}
\label{stab:proportion partial}
\end{table}

\supplementarysection{Diagnostic Reasoning Quality across Medical Specialties}{ssec:suppl_reasoning}

\begin{table}[H]
\centering
\makebox[\linewidth][c]{\resizebox{1.075\linewidth}{!}{%
\begin{tblr}{
  colspec={Q[l,wd=4.15cm] *{8}{Q[c]}},
  rowsep=3.4pt,
  row{1}={bg=Stone1,font=\bfseries},
  column{1}={font=\bfseries,cmd=\mbox},
  hlines = {Stone1},
  vlines = {Stone1}
}
Model & \shortstack{Derm.\\(n=63)} & \shortstack{Emerg.\\(n=23)} & \shortstack{Intern.\\(n=23)} & \shortstack{Neph.\\(n=29)} & \shortstack{Neuro.\\(n=98)} & \shortstack{Onco.\\(n=114)} & \shortstack{Ophth.\\(n=60)} & \shortstack{Radiol.\\(n=679)} \\

Claude-4.5-Sonnet & \shortstack{0.479\\ {[0.419, 0.540]}} & \shortstack{0.590\\ {[0.512, 0.669]}} & \shortstack{0.490\\ {[0.417, 0.563]}} & \shortstack{0.503\\ {[0.430, 0.574]}} & \shortstack{0.639\\ {[0.607, 0.672]}} & \shortstack{0.578\\ {[0.545, 0.612]}} & \shortstack{0.611\\ {[0.565, 0.655]}} & \shortstack{0.512\\ {[0.498, 0.526]}} \\

Gemini 3 Pro & \shortstack{0.475\\ {[0.424, 0.527]}} & \shortstack{0.544\\ {[0.467, 0.621]}} & \shortstack{0.465\\ {[0.418, 0.513]}} & \shortstack{0.515\\ {[0.442, 0.588]}} & \shortstack{0.597\\ {[0.564, 0.630]}} & \shortstack{0.528\\ {[0.496, 0.560]}} & \shortstack{0.585\\ {[0.544, 0.627]}} & \shortstack{0.519\\ {[0.506, 0.532]}} \\

GPT-5-minimal & \shortstack{0.480\\ {[0.423, 0.537]}} & \shortstack{0.540\\ {[0.475, 0.607]}} & \shortstack{0.495\\ {[0.435, 0.554]}} & \shortstack{0.553\\ {[0.485, 0.614]}} & \shortstack{0.678\\ {[0.648, 0.709]}} & \shortstack{0.590\\ {[0.557, 0.622]}} & \shortstack{0.659\\ {[0.618, 0.700]}} & \shortstack{0.570\\ {[0.556, 0.584]}} \\

GPT-5-medium & \shortstack{0.491\\ {[0.429, 0.552]}} & \shortstack{0.590\\ {[0.516, 0.664]}} & \shortstack{0.482\\ {[0.410, 0.555]}} & \shortstack{0.574\\ {[0.497, 0.650]}} & \shortstack{0.682\\ {[0.647, 0.716]}} & \shortstack{0.617\\ {[0.581, 0.652]}} & \shortstack{0.658\\ {[0.616, 0.701]}} & \shortstack{0.594\\ {[0.580, 0.609]}} \\

Gemma-3-4B & \shortstack{0.297\\ {[0.250, 0.347]}} & \shortstack{0.370\\ {[0.295, 0.446]}} & \shortstack{0.306\\ {[0.259, 0.351]}} & \shortstack{0.358\\ {[0.302, 0.415]}} & \shortstack{0.475\\ {[0.443, 0.507]}} & \shortstack{0.405\\ {[0.377, 0.434]}} & \shortstack{0.450\\ {[0.401, 0.499]}} & \shortstack{0.382\\ {[0.370, 0.393]}} \\

Gemma-3-27B & \shortstack{0.383\\ {[0.333, 0.435]}} & \shortstack{0.463\\ {[0.379, 0.549]}} & \shortstack{0.343\\ {[0.287, 0.398]}} & \shortstack{0.408\\ {[0.343, 0.479]}} & \shortstack{0.537\\ {[0.504, 0.571]}} & \shortstack{0.464\\ {[0.429, 0.499]}} & \shortstack{0.534\\ {[0.491, 0.578]}} & \shortstack{0.451\\ {[0.439, 0.464]}} \\

LLaMA-4-Scout & \shortstack{0.352\\ {[0.305, 0.402]}} & \shortstack{0.376\\ {[0.316, 0.439]}} & \shortstack{0.336\\ {[0.284, 0.390]}} & \shortstack{0.395\\ {[0.327, 0.467]}} & \shortstack{0.487\\ {[0.459, 0.517]}} & \shortstack{0.417\\ {[0.385, 0.450]}} & \shortstack{0.464\\ {[0.416, 0.511]}} & \shortstack{0.416\\ {[0.404, 0.427]}} \\

Qwen3-VL-4B & \shortstack{0.383\\ {[0.338, 0.429]}} & \shortstack{0.467\\ {[0.392, 0.542]}} & \shortstack{0.393\\ {[0.343, 0.446]}} & \shortstack{0.417\\ {[0.352, 0.482]}} & \shortstack{0.569\\ {[0.533, 0.604]}} & \shortstack{0.503\\ {[0.472, 0.535]}} & \shortstack{0.543\\ {[0.499, 0.587]}} & \shortstack{0.483\\ {[0.470, 0.495]}} \\

Qwen3-VL-8B & \shortstack{0.380\\ {[0.330, 0.431]}} & \shortstack{0.501\\ {[0.430, 0.575]}} & \shortstack{0.385\\ {[0.321, 0.451]}} & \shortstack{0.417\\ {[0.355, 0.481]}} & \shortstack{0.591\\ {[0.558, 0.624]}} & \shortstack{0.508\\ {[0.476, 0.541]}} & \shortstack{0.573\\ {[0.526, 0.618]}} & \shortstack{0.487\\ {[0.474, 0.500]}} \\

Qwen3-VL-32B & \shortstack{0.438\\ {[0.388, 0.489]}} & \shortstack{0.507\\ {[0.432, 0.585]}} & \shortstack{0.418\\ {[0.337, 0.498]}} & \shortstack{0.468\\ {[0.404, 0.527]}} & \shortstack{0.633\\ {[0.599, 0.666]}} & \shortstack{0.525\\ {[0.490, 0.561]}} & \shortstack{0.597\\ {[0.550, 0.644]}} & \shortstack{0.516\\ {[0.503, 0.530]}} \\

Qwen3-VL-4B-thinking & \shortstack{0.345\\ {[0.299, 0.392]}} & \shortstack{0.409\\ {[0.342, 0.480]}} & \shortstack{0.376\\ {[0.314, 0.440]}} & \shortstack{0.361\\ {[0.316, 0.407]}} & \shortstack{0.567\\ {[0.533, 0.602]}} & \shortstack{0.466\\ {[0.437, 0.496]}} & \shortstack{0.534\\ {[0.489, 0.577]}} & \shortstack{0.446\\ {[0.434, 0.459]}} \\

Qwen3-VL-8B-thinking & \shortstack{0.386\\ {[0.336, 0.436]}} & \shortstack{0.465\\ {[0.393, 0.542]}} & \shortstack{0.416\\ {[0.366, 0.465]}} & \shortstack{0.397\\ {[0.343, 0.452]}} & \shortstack{0.560\\ {[0.527, 0.593]}} & \shortstack{0.472\\ {[0.436, 0.509]}} & \shortstack{0.560\\ {[0.517, 0.602]}} & \shortstack{0.459\\ {[0.447, 0.472]}} \\

Qwen3-VL-32B-thinking & \shortstack{0.419\\ {[0.363, 0.474]}} & \shortstack{0.458\\ {[0.382, 0.536]}} & \shortstack{0.428\\ {[0.358, 0.500]}} & \shortstack{0.468\\ {[0.393, 0.544]}} & \shortstack{0.573\\ {[0.542, 0.604]}} & \shortstack{0.467\\ {[0.433, 0.502]}} & \shortstack{0.520\\ {[0.472, 0.569]}} & \shortstack{0.449\\ {[0.435, 0.462]}} \\

MedGemma-4B & \shortstack{0.277\\ {[0.238, 0.317]}} & \shortstack{0.359\\ {[0.302, 0.418]}} & \shortstack{0.266\\ {[0.214, 0.320]}} & \shortstack{0.310\\ {[0.247, 0.377]}} & \shortstack{0.474\\ {[0.439, 0.510]}} & \shortstack{0.401\\ {[0.372, 0.432]}} & \shortstack{0.443\\ {[0.401, 0.485]}} & \shortstack{0.376\\ {[0.364, 0.387]}} \\

MedGemma-27B & \shortstack{0.379\\ {[0.330, 0.430]}} & \shortstack{0.457\\ {[0.384, 0.532]}} & \shortstack{0.368\\ {[0.297, 0.444]}} & \shortstack{0.387\\ {[0.318, 0.463]}} & \shortstack{0.558\\ {[0.523, 0.594]}} & \shortstack{0.478\\ {[0.445, 0.510]}} & \shortstack{0.529\\ {[0.484, 0.576]}} & \shortstack{0.466\\ {[0.453, 0.479]}} \\

Overall & \shortstack{0.398\\ {[0.354, 0.441]}} & \shortstack{0.473\\ {[0.417, 0.529]}} & \shortstack{0.398\\ {[0.352, 0.444]}} & \shortstack{0.435\\ {[0.381, 0.489]}} & \shortstack{0.575\\ {[0.550, 0.599]}} & \shortstack{0.495\\ {[0.470, 0.520]}} & \shortstack{0.551\\ {[0.516, 0.584]}} & \shortstack{0.475\\ {[0.465, 0.485]}} \\
\end{tblr}
}}
\caption{\textbf{Fact recall scores across medical specialties on
ClinMM-Bench.} Values indicate mean fact recall scores, with 95\%
confidence intervals estimated using 100,000 bootstrap resamples.}
\label{stab:fact recall}
\end{table}

\begin{table}[H]
\centering
\makebox[\linewidth][c]{\resizebox{1.075\linewidth}{!}{%
\begin{tblr}{
  colspec={Q[l,wd=4.15cm] *{8}{Q[c]}},
  rowsep=3.4pt,
  row{1}={bg=Stone1,font=\bfseries},
  column{1}={font=\bfseries,cmd=\mbox},
  hlines = {Stone1},
  vlines = {Stone1}
}
Model & \shortstack{Derm.\\(n=63)} & \shortstack{Emerg.\\(n=23)} & \shortstack{Intern.\\(n=23)} & \shortstack{Neph.\\(n=29)} & \shortstack{Neuro.\\(n=98)} & \shortstack{Onco.\\(n=114)} & \shortstack{Ophth.\\(n=60)} & \shortstack{Radiol.\\(n=679)} \\

Claude-4.5-Sonnet & \shortstack{0.074\\ {[0.052, 0.097]}} & \shortstack{0.108\\ {[0.059, 0.166]}} & \shortstack{0.097\\ {[0.051, 0.150]}} & \shortstack{0.081\\ {[0.050, 0.121]}} & \shortstack{0.060\\ {[0.046, 0.074]}} & \shortstack{0.092\\ {[0.074, 0.112]}} & \shortstack{0.067\\ {[0.045, 0.092]}} & \shortstack{0.142\\ {[0.131, 0.153]}} \\

Gemini 3 Pro & \shortstack{0.060\\ {[0.039, 0.085]}} & \shortstack{0.052\\ {[0.019, 0.093]}} & \shortstack{0.091\\ {[0.050, 0.138]}} & \shortstack{0.080\\ {[0.041, 0.130]}} & \shortstack{0.048\\ {[0.035, 0.062]}} & \shortstack{0.075\\ {[0.055, 0.098]}} & \shortstack{0.059\\ {[0.039, 0.082]}} & \shortstack{0.098\\ {[0.088, 0.109]}} \\

GPT-5-minimal & \shortstack{0.067\\ {[0.046, 0.090]}} & \shortstack{0.070\\ {[0.041, 0.103]}} & \shortstack{0.098\\ {[0.049, 0.161]}} & \shortstack{0.077\\ {[0.046, 0.112]}} & \shortstack{0.069\\ {[0.051, 0.089]}} & \shortstack{0.091\\ {[0.072, 0.111]}} & \shortstack{0.063\\ {[0.041, 0.088]}} & \shortstack{0.132\\ {[0.121, 0.144]}} \\

GPT-5-medium & \shortstack{0.090\\ {[0.063, 0.119]}} & \shortstack{0.091\\ {[0.041, 0.153]}} & \shortstack{0.100\\ {[0.046, 0.166]}} & \shortstack{0.080\\ {[0.049, 0.115]}} & \shortstack{0.056\\ {[0.040, 0.074]}} & \shortstack{0.075\\ {[0.055, 0.096]}} & \shortstack{0.063\\ {[0.038, 0.091]}} & \shortstack{0.124\\ {[0.112, 0.135]}} \\

Gemma-3-4B & \shortstack{0.156\\ {[0.128, 0.184]}} & \shortstack{0.229\\ {[0.165, 0.292]}} & \shortstack{0.186\\ {[0.128, 0.250]}} & \shortstack{0.141\\ {[0.101, 0.186]}} & \shortstack{0.180\\ {[0.155, 0.205]}} & \shortstack{0.205\\ {[0.178, 0.232]}} & \shortstack{0.198\\ {[0.158, 0.240]}} & \shortstack{0.268\\ {[0.254, 0.282]}} \\

Gemma-3-27B & \shortstack{0.131\\ {[0.103, 0.162]}} & \shortstack{0.152\\ {[0.091, 0.220]}} & \shortstack{0.192\\ {[0.130, 0.261]}} & \shortstack{0.107\\ {[0.076, 0.139]}} & \shortstack{0.114\\ {[0.092, 0.137]}} & \shortstack{0.175\\ {[0.148, 0.203]}} & \shortstack{0.123\\ {[0.092, 0.158]}} & \shortstack{0.203\\ {[0.191, 0.215]}} \\

LLaMA-4-Scout & \shortstack{0.079\\ {[0.056, 0.105]}} & \shortstack{0.141\\ {[0.084, 0.204]}} & \shortstack{0.127\\ {[0.082, 0.175]}} & \shortstack{0.082\\ {[0.047, 0.123]}} & \shortstack{0.081\\ {[0.063, 0.099]}} & \shortstack{0.136\\ {[0.113, 0.161]}} & \shortstack{0.099\\ {[0.072, 0.128]}} & \shortstack{0.148\\ {[0.137, 0.160]}} \\

Qwen3-VL-4B & \shortstack{0.117\\ {[0.094, 0.143]}} & \shortstack{0.123\\ {[0.086, 0.163]}} & \shortstack{0.170\\ {[0.121, 0.224]}} & \shortstack{0.096\\ {[0.070, 0.123]}} & \shortstack{0.116\\ {[0.097, 0.137]}} & \shortstack{0.140\\ {[0.122, 0.159]}} & \shortstack{0.130\\ {[0.102, 0.159]}} & \shortstack{0.175\\ {[0.166, 0.185]}} \\

Qwen3-VL-8B & \shortstack{0.113\\ {[0.090, 0.138]}} & \shortstack{0.112\\ {[0.068, 0.159]}} & \shortstack{0.154\\ {[0.109, 0.201]}} & \shortstack{0.131\\ {[0.097, 0.166]}} & \shortstack{0.116\\ {[0.097, 0.137]}} & \shortstack{0.135\\ {[0.117, 0.153]}} & \shortstack{0.112\\ {[0.090, 0.135]}} & \shortstack{0.173\\ {[0.163, 0.184]}} \\

Qwen3-VL-32B & \shortstack{0.116\\ {[0.088, 0.148]}} & \shortstack{0.070\\ {[0.039, 0.108]}} & \shortstack{0.114\\ {[0.068, 0.168]}} & \shortstack{0.095\\ {[0.059, 0.135]}} & \shortstack{0.076\\ {[0.060, 0.093]}} & \shortstack{0.134\\ {[0.112, 0.157]}} & \shortstack{0.102\\ {[0.073, 0.134]}} & \shortstack{0.166\\ {[0.155, 0.177]}} \\

Qwen3-VL-4B-thinking & \shortstack{0.144\\ {[0.116, 0.173]}} & \shortstack{0.142\\ {[0.100, 0.186]}} & \shortstack{0.160\\ {[0.110, 0.214]}} & \shortstack{0.134\\ {[0.094, 0.176]}} & \shortstack{0.140\\ {[0.115, 0.165]}} & \shortstack{0.174\\ {[0.149, 0.200]}} & \shortstack{0.149\\ {[0.122, 0.178]}} & \shortstack{0.205\\ {[0.194, 0.216]}} \\

Qwen3-VL-8B-thinking & \shortstack{0.140\\ {[0.111, 0.170]}} & \shortstack{0.113\\ {[0.063, 0.168]}} & \shortstack{0.155\\ {[0.106, 0.210]}} & \shortstack{0.113\\ {[0.076, 0.156]}} & \shortstack{0.126\\ {[0.104, 0.149]}} & \shortstack{0.149\\ {[0.128, 0.171]}} & \shortstack{0.134\\ {[0.106, 0.163]}} & \shortstack{0.195\\ {[0.184, 0.206]}} \\

Qwen3-VL-32B-thinking & \shortstack{0.132\\ {[0.105, 0.160]}} & \shortstack{0.143\\ {[0.092, 0.197]}} & \shortstack{0.162\\ {[0.108, 0.220]}} & \shortstack{0.098\\ {[0.066, 0.134]}} & \shortstack{0.092\\ {[0.074, 0.112]}} & \shortstack{0.167\\ {[0.144, 0.192]}} & \shortstack{0.130\\ {[0.099, 0.163]}} & \shortstack{0.205\\ {[0.193, 0.216]}} \\

MedGemma-4B & \shortstack{0.125\\ {[0.098, 0.153]}} & \shortstack{0.116\\ {[0.068, 0.171]}} & \shortstack{0.138\\ {[0.087, 0.197]}} & \shortstack{0.091\\ {[0.058, 0.136]}} & \shortstack{0.132\\ {[0.107, 0.157]}} & \shortstack{0.147\\ {[0.127, 0.168]}} & \shortstack{0.133\\ {[0.105, 0.163]}} & \shortstack{0.199\\ {[0.187, 0.211]}} \\

MedGemma-27B & \shortstack{0.086\\ {[0.063, 0.111]}} & \shortstack{0.146\\ {[0.075, 0.230]}} & \shortstack{0.142\\ {[0.072, 0.227]}} & \shortstack{0.100\\ {[0.058, 0.146]}} & \shortstack{0.094\\ {[0.072, 0.118]}} & \shortstack{0.139\\ {[0.116, 0.164]}} & \shortstack{0.101\\ {[0.074, 0.132]}} & \shortstack{0.187\\ {[0.175, 0.200]}} \\

Overall & \shortstack{0.109\\ {[0.092, 0.126]}} & \shortstack{0.120\\ {[0.088, 0.155]}} & \shortstack{0.139\\ {[0.103, 0.180]}} & \shortstack{0.100\\ {[0.079, 0.123]}} & \shortstack{0.100\\ {[0.088, 0.112]}} & \shortstack{0.136\\ {[0.124, 0.148]}} & \shortstack{0.111\\ {[0.093, 0.130]}} & \shortstack{0.175\\ {[0.168, 0.182]}} \\
\end{tblr}
}}
\caption{\textbf{Hallucination scores across medical specialties on
ClinMM-Bench.} Values indicate mean hallucination scores, with 95\%
confidence intervals estimated using 100,000 bootstrap resamples.}
\label{stab:hallucination}
\end{table}

\begin{table}[H]
\centering
\makebox[\linewidth][c]{\resizebox{1.075\linewidth}{!}{%
\begin{tblr}{
  colspec={Q[l,wd=4.15cm] *{8}{Q[c]}},
  rowsep=3.4pt,
  row{1}={bg=Stone1,font=\bfseries},
  column{1}={font=\bfseries,cmd=\mbox},
  hlines = {Stone1},
  vlines = {Stone1}
}
Model & \shortstack{Derm.\\(n=63)} & \shortstack{Emerg.\\(n=23)} & \shortstack{Intern.\\(n=23)} & \shortstack{Neph.\\(n=29)} & \shortstack{Neuro.\\(n=98)} & \shortstack{Onco.\\(n=114)} & \shortstack{Ophth.\\(n=60)} & \shortstack{Radiol.\\(n=679)} \\

Claude-4.5-Sonnet & \shortstack{0.254\\ {[0.218, 0.292]}} & \shortstack{0.263\\ {[0.217, 0.308]}} & \shortstack{0.230\\ {[0.183, 0.282]}} & \shortstack{0.266\\ {[0.224, 0.310]}} & \shortstack{0.386\\ {[0.360, 0.412]}} & \shortstack{0.332\\ {[0.307, 0.357]}} & \shortstack{0.344\\ {[0.308, 0.381]}} & \shortstack{0.293\\ {[0.282, 0.303]}} \\

Gemini 3 Pro & \shortstack{0.352\\ {[0.308, 0.396]}} & \shortstack{0.350\\ {[0.299, 0.401]}} & \shortstack{0.347\\ {[0.292, 0.406]}} & \shortstack{0.366\\ {[0.303, 0.430]}} & \shortstack{0.483\\ {[0.454, 0.512]}} & \shortstack{0.425\\ {[0.393, 0.457]}} & \shortstack{0.447\\ {[0.404, 0.491]}} & \shortstack{0.408\\ {[0.395, 0.421]}} \\

GPT-5-minimal & \shortstack{0.311\\ {[0.266, 0.357]}} & \shortstack{0.334\\ {[0.286, 0.384]}} & \shortstack{0.305\\ {[0.251, 0.362]}} & \shortstack{0.329\\ {[0.277, 0.379]}} & \shortstack{0.443\\ {[0.417, 0.470]}} & \shortstack{0.389\\ {[0.360, 0.419]}} & \shortstack{0.418\\ {[0.380, 0.456]}} & \shortstack{0.371\\ {[0.359, 0.384]}} \\

GPT-5-medium & \shortstack{0.265\\ {[0.225, 0.307]}} & \shortstack{0.309\\ {[0.267, 0.351]}} & \shortstack{0.278\\ {[0.229, 0.328]}} & \shortstack{0.306\\ {[0.254, 0.359]}} & \shortstack{0.398\\ {[0.369, 0.427]}} & \shortstack{0.383\\ {[0.355, 0.410]}} & \shortstack{0.376\\ {[0.336, 0.420]}} & \shortstack{0.337\\ {[0.326, 0.348]}} \\

Gemma-3-4B & \shortstack{0.230\\ {[0.193, 0.269]}} & \shortstack{0.260\\ {[0.206, 0.314]}} & \shortstack{0.237\\ {[0.191, 0.282]}} & \shortstack{0.282\\ {[0.232, 0.333]}} & \shortstack{0.386\\ {[0.359, 0.413]}} & \shortstack{0.344\\ {[0.318, 0.370]}} & \shortstack{0.344\\ {[0.300, 0.390]}} & \shortstack{0.313\\ {[0.302, 0.323]}} \\

Gemma-3-27B & \shortstack{0.279\\ {[0.244, 0.314]}} & \shortstack{0.303\\ {[0.251, 0.353]}} & \shortstack{0.233\\ {[0.176, 0.295]}} & \shortstack{0.282\\ {[0.231, 0.339]}} & \shortstack{0.403\\ {[0.376, 0.430]}} & \shortstack{0.370\\ {[0.342, 0.400]}} & \shortstack{0.380\\ {[0.342, 0.418]}} & \shortstack{0.348\\ {[0.336, 0.359]}} \\

LLaMA-4-Scout & \shortstack{0.341\\ {[0.302, 0.381]}} & \shortstack{0.367\\ {[0.313, 0.420]}} & \shortstack{0.328\\ {[0.280, 0.378]}} & \shortstack{0.347\\ {[0.290, 0.407]}} & \shortstack{0.480\\ {[0.450, 0.510]}} & \shortstack{0.456\\ {[0.425, 0.487]}} & \shortstack{0.449\\ {[0.402, 0.495]}} & \shortstack{0.436\\ {[0.423, 0.449]}} \\

Qwen3-VL-4B & \shortstack{0.236\\ {[0.203, 0.272]}} & \shortstack{0.251\\ {[0.196, 0.309]}} & \shortstack{0.211\\ {[0.175, 0.247]}} & \shortstack{0.234\\ {[0.191, 0.278]}} & \shortstack{0.353\\ {[0.325, 0.381]}} & \shortstack{0.313\\ {[0.290, 0.336]}} & \shortstack{0.283\\ {[0.248, 0.318]}} & \shortstack{0.296\\ {[0.286, 0.307]}} \\

Qwen3-VL-8B & \shortstack{0.237\\ {[0.203, 0.273]}} & \shortstack{0.260\\ {[0.215, 0.308]}} & \shortstack{0.242\\ {[0.196, 0.290]}} & \shortstack{0.202\\ {[0.169, 0.238]}} & \shortstack{0.367\\ {[0.340, 0.395]}} & \shortstack{0.315\\ {[0.293, 0.338]}} & \shortstack{0.335\\ {[0.299, 0.372]}} & \shortstack{0.306\\ {[0.296, 0.316]}} \\

Qwen3-VL-32B & \shortstack{0.264\\ {[0.231, 0.298]}} & \shortstack{0.276\\ {[0.227, 0.325]}} & \shortstack{0.256\\ {[0.199, 0.316]}} & \shortstack{0.283\\ {[0.234, 0.337]}} & \shortstack{0.389\\ {[0.360, 0.419]}} & \shortstack{0.342\\ {[0.316, 0.369]}} & \shortstack{0.354\\ {[0.318, 0.392]}} & \shortstack{0.318\\ {[0.307, 0.329]}} \\

Qwen3-VL-4B-thinking & \shortstack{0.237\\ {[0.202, 0.273]}} & \shortstack{0.241\\ {[0.196, 0.289]}} & \shortstack{0.232\\ {[0.185, 0.279]}} & \shortstack{0.240\\ {[0.204, 0.278]}} & \shortstack{0.377\\ {[0.352, 0.404]}} & \shortstack{0.316\\ {[0.292, 0.341]}} & \shortstack{0.320\\ {[0.282, 0.361]}} & \shortstack{0.293\\ {[0.282, 0.304]}} \\

Qwen3-VL-8B-thinking & \shortstack{0.242\\ {[0.208, 0.278]}} & \shortstack{0.277\\ {[0.214, 0.345]}} & \shortstack{0.209\\ {[0.174, 0.245]}} & \shortstack{0.238\\ {[0.198, 0.282]}} & \shortstack{0.367\\ {[0.340, 0.395]}} & \shortstack{0.297\\ {[0.272, 0.323]}} & \shortstack{0.314\\ {[0.279, 0.352]}} & \shortstack{0.283\\ {[0.273, 0.294]}} \\

Qwen3-VL-32B-thinking & \shortstack{0.254\\ {[0.218, 0.290]}} & \shortstack{0.252\\ {[0.202, 0.303]}} & \shortstack{0.250\\ {[0.190, 0.312]}} & \shortstack{0.270\\ {[0.221, 0.321]}} & \shortstack{0.386\\ {[0.358, 0.415]}} & \shortstack{0.311\\ {[0.285, 0.338]}} & \shortstack{0.304\\ {[0.264, 0.346]}} & \shortstack{0.276\\ {[0.266, 0.286]}} \\

MedGemma-4B & \shortstack{0.261\\ {[0.226, 0.299]}} & \shortstack{0.310\\ {[0.249, 0.374]}} & \shortstack{0.201\\ {[0.164, 0.241]}} & \shortstack{0.238\\ {[0.193, 0.286]}} & \shortstack{0.405\\ {[0.373, 0.438]}} & \shortstack{0.389\\ {[0.362, 0.416]}} & \shortstack{0.377\\ {[0.336, 0.420]}} & \shortstack{0.358\\ {[0.347, 0.369]}} \\

MedGemma-27B & \shortstack{0.259\\ {[0.219, 0.300]}} & \shortstack{0.290\\ {[0.247, 0.333]}} & \shortstack{0.215\\ {[0.155, 0.287]}} & \shortstack{0.253\\ {[0.207, 0.303]}} & \shortstack{0.372\\ {[0.347, 0.399]}} & \shortstack{0.337\\ {[0.309, 0.365]}} & \shortstack{0.334\\ {[0.293, 0.377]}} & \shortstack{0.318\\ {[0.307, 0.329]}} \\

Overall & \shortstack{0.268\\ {[0.239, 0.298]}} & \shortstack{0.290\\ {[0.251, 0.326]}} & \shortstack{0.252\\ {[0.219, 0.286]}} & \shortstack{0.276\\ {[0.240, 0.311]}} & \shortstack{0.400\\ {[0.382, 0.417]}} & \shortstack{0.355\\ {[0.336, 0.373]}} & \shortstack{0.359\\ {[0.330, 0.387]}} & \shortstack{0.330\\ {[0.323, 0.338]}} \\

\end{tblr}
}}
\caption{\textbf{Fact density scores across medical specialties on
ClinMM-Bench.} Values indicate mean fact density scores, with 95\%
confidence intervals estimated using 100,000 bootstrap resamples.}
\label{stab:fact density}
\end{table}

\supplementarysection{Search Strategy}{ssec:search}

Dermatology: (dermatology[Title/Abstract] AND case
reports[Publication Type] AND pubmed pmc open access[Filter] AND
humans[Filter] AND english[Filter])

Emergency Medicine: (emergency medicine[Title/Abstract] AND case
reports[Publication Type] AND pubmed pmc open access[Filter] AND
humans[Filter])

Internal Medicine: (internal medicine[Title/Abstract] AND case
reports[Publication Type] AND pubmed pmc open access[Filter] AND
humans[Filter])

Nephrology: (nephrology[Title/Abstract] AND case reports[Publication
Type] AND pubmed pmc open access[Filter] AND humans[Filter] AND
english[Filter])

Neurology: (neurology[Title/Abstract] AND case reports[Publication
Type] AND pubmed pmc open access[Filter] AND humans[Filter] AND
english[Filter])

Oncology: (oncology[Title/Abstract] AND case reports[Publication Type]
AND pubmed pmc open access[Filter] AND humans[Filter] AND
english[Filter])

Ophthalmology: (ophthalmology[Title/Abstract] AND case
reports[Publication Type] AND pubmed pmc open access[Filter] AND
humans[Filter] AND english[Filter])

Radiology: (radiology[Title/Abstract] AND case reports[Publication
Type] AND pubmed pmc open access[Filter] AND english[Filter])

\supplementarysection{Data Curation Flow}{ssec:suppl_dataflow}

\begin{table}[H]
\centering
\footnotesize
\begin{tblr}{
  colspec={Q[l,m] *{6}{Q[c,m]}},
  rowsep=3.4pt,
  row{1}={bg=Stone1,font=\bfseries,halign=c},
  column{1}={font=\bfseries},
  row{Z}={bg=Stone1,font=\bfseries},
  hlines = {Stone1},
  vlines = {Stone1}
}
Specialty & Collection \& Extraction & Inspection & Validation & Conversion & Quality Control & Included Cases \\

Derm. & 265 & 139 & 68 & 68 & 66 & 63 \\
Emerg. & 129 & 65 & 25 & 25 & 25 & 23 \\
Intern. & 97 & 39 & 24 & 24 & 24 & 23 \\
Neph. & 127 & 66 & 32 & 32 & 32 & 29 \\
Neuro. & 400 & 189 & 101 & 101 & 101 & 98 \\
Onco. & 758 & 373 & 123 & 123 & 122 & 114 \\
Ophth. & 284 & 123 & 62 & 62 & 61 & 60 \\
Radiol. & 2,582 & 1,387 & 709 & 709 & 693 & 679 \\

Total & 4,642 & 2,381 & 1,144 & 1,144 & 1,124 & 1,089 \\

\end{tblr}

\caption{\textbf{Data curation flow across medical specialties.} Counts
are shown after each sequential data-curation stage; the final column
reports cases retained for benchmark evaluation after all curation
stages.}
\label{stab:data}
\end{table}

\supplementarysection{Data Validation Prompt}{prompt:data validation}

\begin{pboxed}
Please determine whether the following medical case report is suitable
for use as a diagnostic teaching case (similar to USMLE format).

\textbf{A case report is suitable as a diagnostic teaching case if it
meets the following criteria:}
\begin{enumerate}
\item Describes a complete diagnostic process for a single patient;
\item Provides relevant clinical information such as demographic
characteristics, chief complaint, medical history, physical examination,
laboratory and imaging findings;
\item Includes key diagnostic images (check for \verb|<fig>|
tags with \verb|<graphic>| elements):
    \begin{itemize}
    \item Initial clinical photographs (e.g., lesions, symptoms);
    \item Diagnostic imaging (e.g., X-ray, CT, MRI, ultrasound, endoscopy);
    \item Other relevant test images (e.g., ECG, pathology slides, blood
smears);
    \end{itemize}
\item States the final diagnosis clearly;
\item Contains the reasoning process that led to the diagnosis.
\end{enumerate}

\textbf{Characteristics of an unsuitable report
(NOT\_DIAGNOSTIC\_SUITABLE):}
\begin{enumerate}
\item Focuses on treatment outcomes, surgical techniques, or follow-up
results;
\item Reports adverse drug reactions or complication management;
\item Consists of literature reviews, theoretical analyses, or opinion
statements;
\item Lacks a diagnostic reasoning process and contains only simple case
descriptions.
\end{enumerate}

\textbf{Case Report:}\\
\verb|{case_report}|

\textbf{Output:}\\
Please respond with only: \verb|"DIAGNOSTIC_SUITABLE"| or
\verb|"NOT_DIAGNOSTIC_SUITABLE"|
\end{pboxed}

\supplementarysection{Data Conversion Prompt}{prompt:data conversion}

\begin{pboxed}[enlargepage flexible=2\baselineskip]
You are a medical information extraction expert. Your task is to convert
a medical case report into a structured JSON format. Return only the
JSON result. Do not include explanations or additional comments.

\textbf{Task}

Convert the original medical case report into the following JSON
format:

\begin{lstlisting}[language=json]
{
 "case_presentation": {
   "english": {
     "demographic_information": "",
     "chief_complaint": "",
     "history_of_present_illness": "",
     "lab_result": "",
     "physical_examination": ""
   },
   "chinese": {
     "demographic_information": "",
     "chief_complaint": "",
     "history_of_present_illness": "",
     "lab_result": "",
     "physical_examination": ""
   }
 },
 "images": {
   "english": [
     {
       "image_id": "",
       "image_type": "",
       "image_finding": "",
       "diagnostic_significance": ""
     }
   ],
   "chinese": [
     {
       "image_id": "",
       "image_type": "",
       "image_finding": "",
       "diagnostic_significance": ""
     }
   ]
 },
 "final_diagnosis": {
   "english": "",
   "chinese": ""
 },
 "diagnostic_reasoning": {
   "english": "",
   "chinese": ""
 }
}
\end{lstlisting}

\textbf{Conversion Rules}

\textbf{Case Presentation MUST NOT contain:}
\begin{enumerate}[nosep,leftmargin=1.5em]
\item Final Diagnosis:
- Case presentation must absolutely not reveal the final diagnosis.
\item Image Diagnostic Information:
    \begin{itemize}
    \item Any information that can only be obtained through image analysis must not appear in the case presentation, including but not limited to:
    \item Specific results of imaging examinations.
    \item Pathological examination cell morphology and staining results.
    \item Endoscopic examination lesion characteristics.
    \item Dermatological lesion morphological descriptions.
    \end{itemize}
\end{enumerate}

\textbf{Case Presentation CAN include:}
\begin{itemize}
\item Patient information (age, gender, etc.).
\item Symptom descriptions.
\item Medical history.
\item Laboratory test values (blood routine, biochemistry, etc.).
\item Examination process descriptions (e.g., ``CT scan performed", ``biopsy taken").
\end{itemize}

\textbf{Physical Examination Field:}\\
Only write what examinations were performed, NOT the examination
results.

Acceptable Examples:
\begin{itemize}
\item ``Chest CT scan performed".
\item ``Bone marrow biopsy conducted".
\item ``Immunohistochemical staining performed".
\item ``PET scan examination conducted".
\end{itemize}

Not Allowed:
\begin{itemize}
\item ``CT showed bilateral lung lesions".
\item ``Biopsy revealed malignant cells".
\end{itemize}

\textbf{Images Field:}\\
For each image/figure mentioned in the original text, extract:
\begin{itemize}
\item \verb|image_id|: The figure number/label from the original text.
\item \verb|image_type|: Concise description of the image type (e.g., ``H\&E staining histological image", ``Chest CT scan", ``Dermatological photograph", ``Immunohistochemistry stain").
\item \verb|image_finding|: What should be visible in this image based on the original text (e.g., ``Malignant cells with nuclear atypia", ``Bilateral lung nodules", ``Erythematous skin lesions").
\item \verb|diagnostic_significance|: What this image contributes to the diagnosis (e.g., ``Confirms malignancy", ``Shows metastatic spread", ``Indicates inflammatory process").
\end{itemize}

\textbf{Final Diagnosis Field:}\\
Write out the clear final diagnosis as stated in the original text.

\textbf{Diagnostic Reasoning Field:}\\
Explain how the diagnosis was reached based on the diagnostic evidence and process described in the original text. Include:
\begin{itemize}
\item How clinical findings contributed to the diagnosis
\item How each image/test result supported or ruled out differential diagnoses
\item The logical progression from initial presentation to final diagnosis
\end{itemize}

\textbf{Must be strictly based on original text content, cannot add any content not in the original:}
\begin{itemize}
\item Cannot perform your own medical reasoning or add medical knowledge.
\item Cannot speculate or supplement information not mentioned in the
original text.
\end{itemize}

\textbf{Case Report:}\\
\verb|{case_report}|

\textbf{Output}\\
Please return ONLY the JSON format result:
\end{pboxed}

\supplementarysection{Quality Control Prompt}{prompt:data quality}

\begin{pboxed}
You are a medical expert. Your task is to evaluate the quality of a
converted medical case report for use as a diagnostic examination case.
The converted content must strictly adhere to the facts of the original
case report, without speculation or additions.

\textbf{You will be provided with:}
\begin{enumerate}
\item Original XML medical case report
\item Converted JSON structured format
\end{enumerate}

Please evaluate the conversion quality.

\textbf{Evaluation Criteria}

\textbf{Information Accuracy Score (1-5 Scale)}

Rate how accurately the converted JSON preserves information from the
original XML:\\
\textbf{5 - Excellent}: All information perfectly extracted and preserved\\
\textbf{4 - Good}: Minor discrepancies that don\textquotesingle t affect
meaning\\
\textbf{3 - Acceptable}: Some information loss but key elements preserved\\
\textbf{2 - Poor}: Significant information loss or errors\\
\textbf{1 - Unacceptable}: Major information loss or errors

\textbf{Original XML Case Report:}\\
\verb|{original_xml}|

\textbf{Converted JSON Case Report:}\\
\verb|{converted_json}|

\textbf{Output Format}\\
Please provide your evaluation in the following JSON format:
\begin{lstlisting}[language=json]
{
  "information_accuracy_score": 1-5,
  "information_accuracy_reason": "Explanation of accuracy assessment"
}
\end{lstlisting}

Return only the JSON result without additional explanations.\strut
\end{pboxed}

\supplementarysection{Diagnostic Accuracy Evaluation Prompt}{prompt:diagnostic}

\begin{pboxed}
Please determine whether the predicted diagnosis is correct.

\textbf{Predicted Diagnosis:}\\
\verb|{predicted_diagnosis}|

\textbf{Ground Truth Diagnosis:}\\
\verb|{ground_truth_diagnosis}|

\textbf{Accuracy Evaluation:}\\
0 - Completely Incorrect\\
1 - Partially Correct\\
2 - Completely Correct

Output ONLY the following valid XML (no other text):

\begin{lstlisting}[language=xml]
<result>
  <reasoning>your detailed explanation here</reasoning>
  <accuracy>0 or 1 or 2</accuracy>
</result>
\end{lstlisting}
\end{pboxed}

\supplementarysection{I1. Atomic fact extraction prompt}{prompt:fact extraction}

\begin{pboxed}
You are a medical expert. Extract atomic facts from the diagnostic
reasoning text.

\textbf{An atomic fact should:}
\begin{enumerate}
\item Be a single, indivisible piece of medical information
\item Contain one subject-predicate-object relationship
\item Be specific and factual (avoid vague statements)
\end{enumerate}

Examples of good atomic facts:
\begin{itemize}
\item ``RPR test is positive"
\item ``CSF shows pleocytosis"
\item ``Fundus exam reveals vitritis"
\end{itemize}

Examples of what to avoid:
\begin{itemize}
\item ``Patient has multiple symptoms" (too vague)
\item ``Tests were done" (not specific enough)
\end{itemize}

Return only valid JSON in this exact format:

\begin{lstlisting}[language=json]
{
  "facts": [
    "atomic fact 1",
    "atomic fact 2"
  ]
}
\end{lstlisting}
\end{pboxed}

\clearpage

\subsection*{I2. Atomic Fact Matching Prompt}
\label{prompt:fact matching}

\begin{pboxed}
You are a medical expert. Compare predicted atomic facts against
ground-truth atomic facts.

\textbf{Tasks:}
\begin{enumerate}
\item Match facts that express the same clinical meaning.
\item Allow N-to-M matching: multiple predicted facts may match one
ground-truth fact, and one predicted fact may match multiple
ground-truth facts.
\item Identify hallucinated predicted facts: predicted facts that are
factually incorrect or contradict the ground truth.
\end{enumerate}

\textbf{Clinical Matching Rules:}
\begin{itemize}
\item Do not match facts if the specimen differs, such as serum vs CSF.
\item Do not match facts if the anatomical location or laterality differs.
\item Do not match positive findings with negative, suspected, or ruled-out
findings.
\item Do not match causal or diagnostic relationships if the direction is
different.
\item If an unmatched predicted fact is merely not covered by the ground
truth but is not clearly wrong, do not label it as hallucinated.
\end{itemize}

\textbf{Important Output Rules:}
\begin{itemize}
\item Use IDs only. Do not copy fact text into the JSON.
\item Predicted IDs must come from the P-list.
\item Ground-truth IDs must come from the G-list.
\item Every unmatched predicted ID must appear in exactly one of:
\verb|hallucinated_predicted_ids| or
\verb|non_hallucinated_unmatched_predicted_ids|.
\item A predicted ID must not appear in both \verb|matched_pairs| and either
\verb|hallucinated_predicted_ids| or
\verb|non_hallucinated_unmatched_predicted_ids|.
\end{itemize}

Return only valid JSON in this exact format:
\begin{lstlisting}[language=json]
{
  "matched_pairs": [
    {
      "reasoning": "short explanation",
      "predicted_ids": [1],
      "ground_truth_ids": [1, 2]
    }
  ],
  "hallucinated_predicted_ids": [3],
  "non_hallucinated_unmatched_predicted_ids": [5],
  "unmatched_ground_truth_ids": [4]
}
\end{lstlisting}
\textbf{Predicted Facts:}\\
\verb|{predicted_facts}|

\textbf{Ground-Truth Facts:}\\
\verb|{ground_truth_facts}|
\end{pboxed}

\end{document}